\documentclass[10pt,twocolumn,letterpaper]{article}

\usepackage{array}
\usepackage{colortbl}
\usepackage{multirow}
\usepackage{booktabs}
\usepackage[export]{adjustbox}
\usepackage{arydshln} 
\usepackage[table]{xcolor}

\definecolor{headergray}{RGB}{242,244,247}
\definecolor{accentblue}{RGB}{235,243,252}
\definecolor{accentpurple}{RGB}{233,232,255}
\definecolor{accenttext}{RGB}{35,80,130}

\usepackage[pagenumbers]{cvpr} 

\definecolor{cvprblue}{rgb}{0.21,0.49,0.74}
\usepackage[pagebackref,breaklinks,colorlinks,allcolors=cvprblue]{hyperref}

\def\confName{CVPR}
\def\confYear{2027}

\title{What Makes a Gavagai? \\Learning Visual Concepts from One Positive and Few Negatives}

\author{Katsuya Ogata\\
The University of Osaka\\
{\tt\small ogata@im.sanken.osaka-u.ac.jp}
\and
Kazuya Nishimura\\
The University of Osaka\\
{\tt\small kazuya.nishimura@is.ids.osaka-u.ac.jp}
\and
Yuta Nakashima\\
The University of Osaka\\
{\tt\small n-yuta@im.sanken.osaka-u.ac.jp}
}

\begin{document}
\maketitle
\begin{abstract}
Humans can recognize novel concepts from only a few
observations by identifying defining visual attributes and
eliminating alternative interpretations.
However, a single observation is consistent with multiple
possible concepts, and accurate visual perception alone
is insufficient to determine the intended interpretation.
Negative evidence can reduce this ambiguity by eliminating
competing hypotheses.
Inspired by Quine's gavagai problem, we introduce the
Gavagai Benchmark to evaluate whether vision-language models
can identify the attributes defining a novel concept
from positive and negative examples.
The benchmark consists of synthetic images depicting novel
combinations of visual attributes and real images of newly
described species.
Using these datasets, we examine how models use negative
examples to identify novel concepts and analyze the
respective roles of visual perception and reasoning.
Several models struggle even when correct attributes are
provided, while others show large gaps when ground-truth
attributes are replaced with model predictions. These
diagnostic comparisons highlight challenges in both attribute
perception and inference from limited evidence.
\end{abstract}    
\section{Introduction}
\label{sec:intro}

\emph{Can vision-language models (VLMs) learn novel visual concepts from only a few observations, as humans do?}~\cite{Zeithamova8259}
This ability is essential for recognizing concepts that cannot be predefined or extensively annotated, such as newly discovered species, rare visual findings, and unfamiliar objects encountered by robots.

\begin{figure}[t]
    \centering
    \includegraphics[width=\columnwidth]{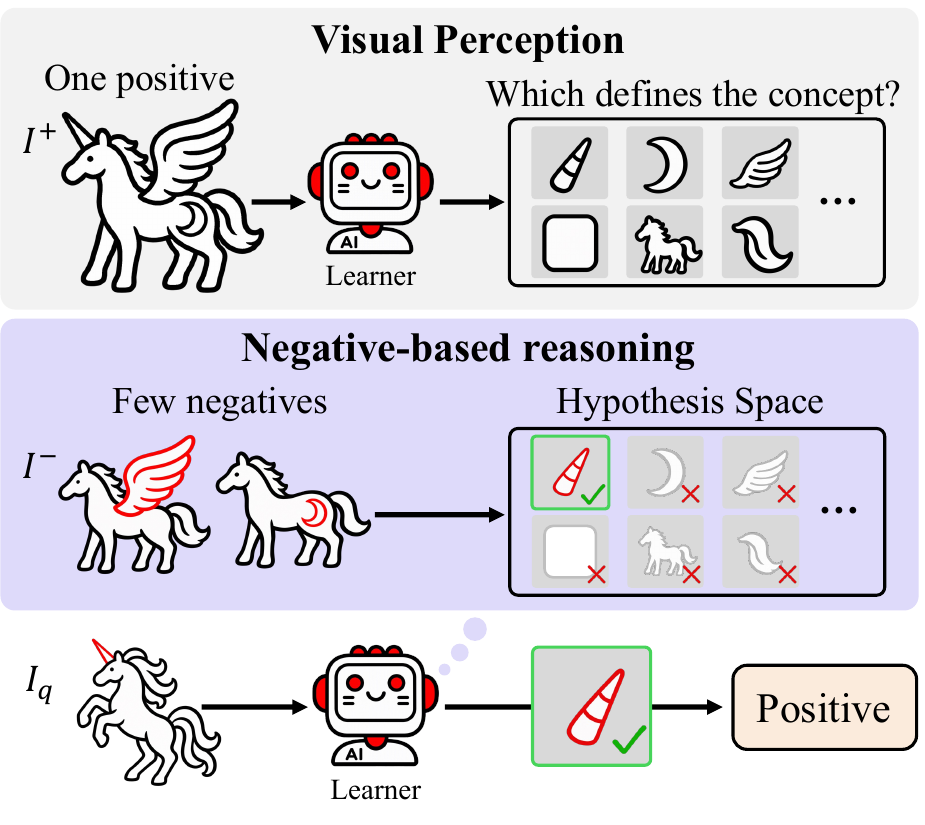}
    \caption{\textbf{Novel visual concept recognition through negative-based reasoning.}
    A single positive example supports multiple hypotheses about the target concept.
    Negative examples eliminate hypotheses that are also consistent with non-target observations, narrowing the possibilities used to classify an unseen query.}
    \label{fig:teaser}
\end{figure}

A fundamental challenge in learning novel visual concepts is their inherent \emph{ambiguity}: a single positive example can support multiple interpretations.
For instance, given a white horse with a horn, wings, and a crescent symbol in \cref{fig:teaser}, the target concept could be defined by its color, shape, symbol, or some combination of these attributes.
This ambiguity echoes Quine's Gavagai problem~\cite{Quine1960}, which illustrates that an observation alone cannot uniquely determine the meaning of a concept.
Even if all visual attributes are correctly perceived, it remains unclear which of them define the target concept.

\emph{Negative examples} provide additional evidence to resolve this ambiguity.
This perspective is based on concept learning theories where learners create and test rules for categories~\cite{Bruner1986,Goodman2008}.
Given a positive example, a learner can form candidate hypotheses about the attributes defining the target concept.
Negative examples then rule out hypotheses that would incorrectly classify them as positive.
By progressively eliminating inconsistent hypotheses, the learner can narrow the possible interpretations and use the remaining hypotheses to classify unseen examples.
We call this process \emph{negative-based reasoning}.
Successful concept recognition thus involves both \emph{visual perception}, which identifies candidate attributes, and \emph{reasoning}, which uses negative evidence to distinguish among competing hypotheses.

Learning visual concepts from limited observations has been studied extensively in few-shot and visual in-context learning.
Bongard-style problems~\cite{bongard,bongardhoi,bongardopenworld} further explore visual rule induction from positive and negative examples, while CURI~\cite{curi} investigates concept learning under uncertainty.
Building on these research directions, we study how VLMs recognize novel visual concepts through negative-based reasoning, with particular emphasis on the roles of visual perception, hypothesis formation, and hypothesis elimination.

To this end, we introduce the \emph{Gavagai Benchmark}, which evaluates novel visual concept recognition from one positive example and a few negative examples.
A model must infer which visual attributes distinguish the target concept from the negatives and apply the inferred concept to classify unseen query images.
Importantly, the task does not necessarily require identifying a unique concept: a query can be classified as long as the remaining hypotheses agree on its label.
This formulation enables us to systematically investigate how VLMs form and eliminate visual hypotheses using negative evidence.

We construct two complementary datasets.
The \emph{Synthetic Gavagai Dataset} consists of generated images with controlled shapes, textures, and patterns.
By systematically varying the defining attribute and the negative examples, we evaluate how models eliminate competing hypotheses and integrate evidence across observations.
Furthermore, providing ground-truth visual attributes or the target rule allows us to distinguish failures in perception from those in reasoning.
The \emph{Real Gavagai Dataset} consists of images of newly described species, whose defining visual attributes are not predefined.
We vary the taxonomic relationships between positive and negative examples to evaluate concept recognition under increasingly fine-grained visual distinctions. Our experiments reveal four key findings.

\noindent\textbf{\includegraphics[height=1.3em,valign=t]{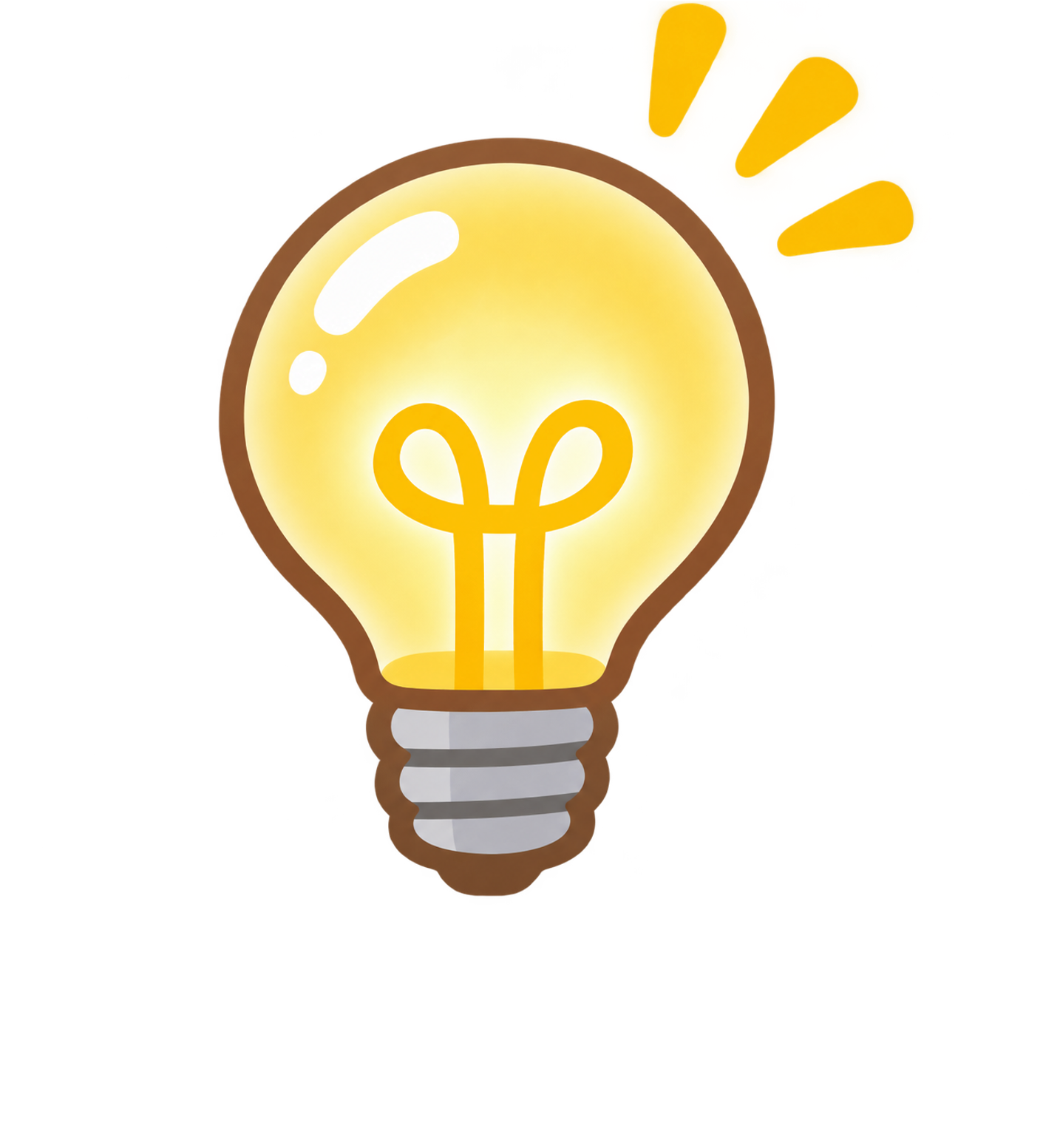} Identifying the relevant hypothesis space is challenging.} Even frontier VLMs, including GPT~\cite{openai2026gpt61terra} and Gemini~\cite{google2026gemini4argon}, benefit substantially from being told which visual attribute categories to consider.

\begin{figure}[t]
  \centering
  \includegraphics[width=\columnwidth]{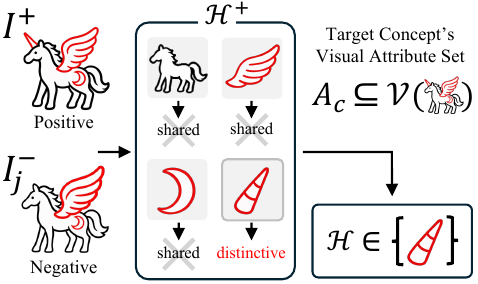}
  \caption{\textbf{Illustration of the problem formulation.} Given a positive example \(I^+\), we consider candidate visual attributes \(\mathcal{H}^+\) that may characterize the target concept. By comparing it with a negative example \(I^-\), shared attributes are eliminated, yielding a refined attribute set \(A_c\). Here, the body stripe pattern is identified as the distinctive attribute of the target concept.}
  \label{fig:formulation}
\end{figure}
\noindent
\textbf{\includegraphics[height=1.3em,valign=t]{fig/bulb.png} Visual similarity alone is insufficient.}
Vision encoder baselines struggle to recognize concepts defined by negative evidence, particularly when the defining attributes are texture or pattern.

\noindent\textbf{\includegraphics[height=1.3em,valign=t]{fig/bulb.png} Reasoning remains challenging even with accurate perception.}
Many VLMs fail to infer the defining attribute even when ground-truth visual attributes are provided, particularly when evidence must be integrated across multiple negatives.

\noindent\textbf{\includegraphics[height=1.3em,valign=t]{fig/bulb.png} Fine-grained recognition is constrained by visual perception.}
Distinguishing newly described species from closely related species remains difficult, and explicit reasoning guidance provides limited improvements at the finest taxonomic level.

These findings reveal distinct challenges in visual perception and negative-based reasoning, showing the importance of both capabilities for visual concept recognition.

\section{Related Work}
\label{sec:related}

\noindent
\textbf{Visual Concept Inference from Few Examples.}
Inferring the meaning of a novel concept from ambiguous observations is a longstanding problem in cognitive science~\cite{Quine1960,Zeithamova8259}. In computer vision, Bongard problems and related benchmarks evaluate whether models can infer visual rules from positive and negative examples~\cite{bongard,bongardlogo,bongardhoi,bongardopenworld}, a capability that remains challenging for current VLMs~\cite{bongardinwonderland}. CURI~\cite{curi} similarly studies concept learning under uncertainty, highlighting the role of negative evidence in resolving competing hypotheses.
Building on this line of work, we consider a setting motivated by real-world encounters with novel concepts, where only a single positive example is available. We examine how VLMs use negative evidence to resolve ambiguity about which visual attributes define the concept.


\noindent
\textbf{Few-Shot and In-Context Visual Learning.}
Few-shot and in-context learning classify new examples by leveraging labeled support examples provided during training or inference~\cite{prototype,few-shot-learning,few-shot-learner,visual-incontext-learning,vl-icl-bench,jiang2024mantis,wang2024muirbench}. Context-Aware Meta-Learning~\cite{fifty2024caml} models support--query relationships for classification, while Link-Context Learning~\cite{tai2024link} strengthens associations between unfamiliar images and novel labels. These approaches primarily focus on support--query matching, rather than reasoning about which visual attributes define the concept.




\noindent
\textbf{Open-Vocabulary Benchmarks for VLMs.}
Open-vocabulary recognition generalizes to unseen categories but typically relies on category names or descriptions~\cite{clip,minderer2022simple,gu2021open,gupta2019lvis,van2018inaturalist,van2021benchmarking,stevens2024bioclip,gu2025bioclip2,parashar2024neglected}. Other work studies novel concepts without prior lexical knowledge, including personalized concept binding~\cite{gal2022image,alaluf2024myvlm} and NVRD~\cite{atur2025nvrd}, which associates unfamiliar visual concepts with nonce words. Related studies also examine novel species beyond known taxonomies~\cite{wang2026deeptaxon,allabadi2025novelspecies,chiranjeevi2025terraincognita}.

\section{Problem Formulation}
\label{sec:problem_formulation}

Inspired by the Gavagai problem \cite{Quine1960}, we introduce a visual concept recognition task in which a target concept must be inferred from a single positive example and a few negative examples.
The challenge is that a positive example contains many visual attributes, and it is unclear which combination of them defines the target concept.
Negative examples help resolve this ambiguity by ruling out possible concepts.

We represent a target concept $c$ as a nonempty set of visual attributes $\mathcal{A}_c$.
Here, an attribute refers to a specific visual property (\eg, red), rather than an attribute category (\eg, color).
Let $\mathcal{V}(I)$ denote the set of attributes present in an image $I$.
We assume that $I$ contains $c$ if and only if all attributes defining $c$ are present in $I$, \ie, $\mathcal{A}_c \subseteq \mathcal{V}(I)$.

Given a positive example $I^+$, any nonempty subset of $\mathcal{V}(I^+)$ could represent the target concept. These subsets span the hypothesis space \begin{align}
    \mathcal{H}^+ =
    \left\{
    h \subseteq \mathcal{V}(I^+)
    \;\middle|\;
    h \neq \emptyset\right\}.
\end{align}

However, a valid concept must not be present in any negative example $I_j^-$.
We can therefore rule out any candidate whose attributes are all present in a negative example.
For instance, if both a positive and a negative example contain the attribute ``red,'' this attribute alone cannot define the target concept, although it may still be part of the concept.
Given a support set
$\mathcal{S} = \{I^+\} \cup \{I_j^-\}_{j=1}^{J}$,
the remaining hypotheses are
\begin{align}
    \mathcal{H} =
    \left\{
    h \in \mathcal{H}^+
    \;\middle|\;
    h \nsubseteq \mathcal{V}(I_j^-)
    \;\forall j
    \right\}.
    \label{eq:negative_elimination}
\end{align}

Our task is to determine whether a query image $I_q$ contains the target concept.
Ideally, we would check whether $\mathcal{A}_c \subseteq \mathcal{V}(I_q)$, but the true concept $\mathcal{A}_c$ is unknown.
Instead, we consider the predictions of all remaining hypotheses.
Each hypothesis $h \in \mathcal{H}$ predicts
$y_q(h) = \mathbb{I}[h \subseteq \mathcal{V}(I_q)]$.
If all hypotheses agree, the query can be classified as positive or negative without knowing the exact target concept.
Thus, we say that $\mathcal{S}$ is sufficient for a query set
$\mathcal{Q} = \{I_q\}_{q=1}^{Q}$ if $y_q(h)$ is the same for all $h \in \mathcal{H}$ and for all $I_q \in \mathcal{Q}$.

Unlike conventional few-shot classification and Bongard problems, our benchmark focuses on reasoning over possible concept definitions by eliminating hypotheses inconsistent with negative evidence.
Importantly, the task does not require uniquely identifying the target concept, as long as the remaining hypotheses agree on the query predictions.

\begin{figure*}[t]
  \centering
  \includegraphics[width=\linewidth]{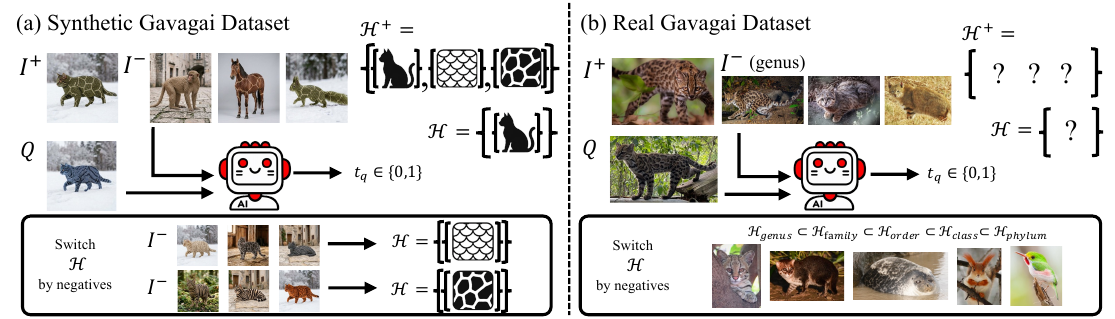}
    \caption{\textbf{Overview of the Synthetic and Real Gavagai datasets.} (a) Synthetic episodes use controlled visual attributes; under a single-attribute prior, negative examples determine the target. (b) Real episodes use recently described species with unknown defining attributes, without a guarantee of identifiability. Both settings classify queries from one positive and multiple negatives.}
  \label{fig:episode_overview}
\end{figure*}

\subsection{Dataset construction}
To evaluate our reasoning-based classification, we construct two datasets: the Synthetic Gavagai Dataset and the Real Gavagai Dataset. 
The Synthetic Gavagai Dataset is designed to assess models' ability to reason over controlled visual attributes by eliminating hypotheses inconsistent with negative examples. The Real Gavagai dataset extends this evaluation to real-world scenarios, assessing the applicability of reasoning-based classification to previously unseen species with distinctive visual characteristics.

\subsubsection{Synthetic Gavagai Dataset Construction}
\label{sec:dataset}

We construct a synthetic dataset with controlled visual attributes to evaluate VLMs' ability to reason by eliminating hypotheses inconsistent with negative examples.

\noindent
\textbf{Visual Attribute Sets.}
We construct a synthetic dataset with three controlled visual factors: shape, texture, and pattern, corresponding to attribute sets $\mathcal{A}_{\mathrm{sha}}$, $\mathcal{A}_{\mathrm{tex}}$, and $\mathcal{A}_{\mathrm{pat}}$.
For example, the positive image $I^+$ in Fig.~\ref{fig:episode_overview}~(a) combines \includegraphics[height=1em,valign=c]{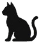}
shape, \includegraphics[height=1em,valign=c]{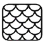} texture, and \includegraphics[height=1em,valign=c]{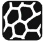} pattern.

\noindent
\textbf{Image Generation.}
We generate synthetic images using the image generation model Gemini 3.1 Flash Image. To ensure that the specified shape, texture, and pattern are reflected in the generated images, we provide an original animal image representing the target shape and a reference image combining the desired texture and pattern. These images are used to guide the generation of composite images. 
We manually verify whether the generated images accurately reflect the specified shape, texture, and pattern, retaining only those that satisfy all three attributes. Of the 1,600 generated images, 1,103 pass manual verification, resulting in a synthetic dataset with controlled visual attributes.

\noindent
\textbf{Episode Construction.}
While the general formulation allows concepts defined
by combinations of attributes, we adopt a
single-attribute hypothesis prior for the synthetic
benchmark:
\begin{align}
    \mathcal{H}_{\mathrm{single}}^+
    =
    \{\{\alpha\} \mid \alpha \in \mathcal{V}(I^+)\}.
\end{align}
We construct negative examples that eliminate all but
one hypothesis under this prior, ensuring that the
support set is sufficient for the query set.
For example, in Fig.~\ref{fig:episode_overview}~(a),
the negatives rule out texture and pattern, leaving
the cat-like shape as the discriminative attribute
($\mathcal{H} = \{\{\text{\includegraphics[height=1em,valign=c]{fig/cat_icon.pdf}}\}\}$).
By varying the negative examples while keeping the
positive image and query set fixed, we construct three
conditions in which shape, texture, or pattern defines
the target concept.

Ground-truth query labels are defined directly by the target
attribute $a$: a query is positive if and only if $a\in\mathcal{V}(I_q)$,
independently of whether $\mathcal{S}$ suffices to infer its label.
evaluation of shape, texture, and pattern.
Each episode contains two positive and six negative
queries, yielding 1,809 episodes in total.
Further details on synthetic dataset construction are
provided in Sec.~\ref{sec:supp-synthetic-dataset-detail}
of the supplementary material.

\subsubsection{Real Gavagai Dataset Construction}
\label{sec:real_dataset}
To evaluate novel concept recognition in real-world settings, we construct a dataset of recently described species whose formal descriptions are unlikely to appear widely in VLM pretraining data. Unlike the synthetic benchmark, the real-world setting does not guarantee support-set sufficiency, as visual attributes and the underlying hypothesis space cannot be explicitly controlled.

\noindent
\textbf{Visual Attribute Sets.}
Unlike synthetic images, natural images contain diverse visual attributes that cannot be explicitly controlled.
To examine concept identification under different levels of visual similarity between positive and negative examples, we leverage the hierarchical structure of biological taxonomy.
As illustrated in Fig.~\ref{fig:episode_overview}~(b),
we consider five taxonomic ranks
$r\in\{\text{phylum},\text{class},\text{order},\text{family},\text{genus}\}$.
Organisms sharing a finer taxonomic rank are expected,
on average, to exhibit greater visual similarity.
We denote by $\mathcal{H}_r$ the hypotheses remaining after
applying Eq.~\eqref{eq:negative_elimination} to the negative
examples selected at rank $r$.
Varying $r$ therefore changes which hypotheses remain consistent
with the support set, while keeping the target species fixed.
More closely related negatives are expected to rule out
hypotheses based on shared visual attributes, requiring
finer distinctions to recognize the target.

\noindent
\textbf{Data Collection.}
For positive examples, we collected 2,418 articles
describing new species published since 2025 from Novataxa\footnote{\url{https://novataxa.blogspot.com/}}.
For negative examples, we retrieved up to 10 research-grade
observation images per taxonomic rank from iNaturalist,
selecting clear images and cropping individual organisms.
The final dataset comprises 297 newly described species,
including 285 animals and 12 fungi.
Each species has one positive example and two to four
positive queries.

\noindent
\textbf{Episode Construction.}
For each target species $c$, we construct five conditions based on the taxonomic rank $r$ defined above.
Each support set contains one positive image and three taxonomically related negative images at rank $r$, as illustrated in Fig.~\ref{fig:episode_overview}~(b). 
Positive queries depict the target species, whereas negative queries depict other species from the same taxonomic group at rank $r$.
Each episode contains two to four queries per class.
In total, we construct 1,480 episodes.
Further details on the real dataset construction are provided in Sec. \ref{sec:supp-real-dataset-detail} of the supplementary material.

\section{Baselines}
\label{sec:method}

We evaluate how effectively VLMs identify the target concept $c$ from the support set $\mathcal{S}$ through in-context inference.
In addition, we investigate how explicitly providing structural priors from our problem formulation (Sec.~\ref{sec:problem_formulation}) affects model performance.

\noindent
\textbf{Common Protocol.}
\label{sec:input}
For each query image $I_q$, the input to a VLM $M$ is
$x_q=(T,\mathcal{S},I_q)$, where $T$ is a text instruction and
$\mathcal{S}$ is the support set. The predicted label
$\hat{t}_q\in\{0,1\}$ is extracted from the response $M(x_q)$.
Each query is evaluated independently to prevent information from
leaking across queries. 

\noindent
\textbf{Structural Priors.}
\label{sec:prompts}
To investigate how prior knowledge affects classification
performance, we evaluate four instruction settings
that progressively introduce task-specific information: \texttt{bare}, \texttt{rule}, \texttt{axes}, and  \texttt{values}.

The \texttt{bare} setting serves as the baseline,
providing only positive and negative labels.
The \texttt{rule} setting additionally specifies that
the target concept is defined by a discriminative
attribute subset $h \subseteq \mathcal{V}(I^+)$,
and negative examples are used to eliminate
inconsistent attribute hypotheses.
The \texttt{axes} setting further reveals that
visual attributes are organized into three categories
(shape, texture, and pattern), and that the target
concept is defined by a single attribute
$h = \{a\}$.
Finally, \texttt{values} provides the complete
vocabulary of candidate attributes in each category.
In the synthetic benchmark, neither \texttt{bare} nor
\texttt{rule} explicitly specifies the single-attribute prior.
These settings therefore evaluate inference of both an
appropriate hypothesis class and the target attribute.

\noindent
\textbf{Reasoning Guidance.}
\label{sec:hea}
Beyond structural priors, we investigate whether explicitly guiding the reasoning process improves concept identification.
We consider two approaches: \texttt{CoT}, which structures reasoning within a single model call, and Hypothesize--Eliminate--Apply (\texttt{HEA}), which decomposes concept acquisition into multiple stages.

In \texttt{CoT}, the model is instructed to describe the support and query images, infer the target concept by comparing positive and negative examples, eliminate candidate attributes shared with negatives, and then output the prediction.
Thus, \texttt{CoT} guides the reasoning process without revealing the predefined attribute categories or values.

\texttt{HEA} explicitly separates reasoning into three stages: \textit{Hypothesize} generates candidate attributes from the positive example, \textit{Eliminate} removes hypotheses inconsistent with the negatives according to Eq.~\eqref{eq:negative_elimination}, and \textit{Apply} classifies the query using the remaining hypotheses.
Each stage is performed as a separate call to the same VLM, allowing intermediate hypotheses to be refined.
\texttt{HEA} follows prior work on hypothesis generation and refinement~\cite{wang2024hypothesis,qiu2024phenomenal} and self-verification~\cite{weng2023large}.
Both \texttt{CoT} and \texttt{HEA} receive the same task information as \texttt{bare}, without access to predefined attribute categories or values.
Detailed prompts are provided in Sec.~\ref{sec:supp-prompts} of the supplementary material.

\section{Experiments and Results}
\label{sec:experiment}

We investigate whether VLMs can identify a novel concept from a single positive example by using negative examples.
On the \textbf{Synthetic Gavagai Dataset}, we first measure how well current VLMs solve the benchmark, and then examine whether task knowledge is required by controlling the prior given in the prompt.
Using controlled inputs, we further analyze the perception and negative-based reasoning abilities of VLMs separately.
We then turn to the \textbf{Real Gavagai Dataset}, which approximates identifying a truly novel species, and test whether VLMs can exploit negative examples in this realistic setting.

\noindent
\textbf{Comparisons.}
We compare three groups of models:
(i)~\emph{vision encoders}: frozen CLIP~\cite{clip}, DINOv2-L~\cite{dinov2}, and SigLIP~\cite{siglip}, which test whether episodes can be solved by visual similarity alone.
The encoders predict $I_q$ as positive if $\cos(I_q, I^+) > \frac{1}{J}\sum_{j=1}^{J}\cos(I_q, I^-_j)$ on $\ell_2$-normalized embeddings, a threshold-free rule that, like the VLMs, uses both positive and negative examples;
(ii)~\emph{commercial VLMs}: GPT-6.1 Terra~\cite{openai2026gpt61terra} and Gemini 4 Argon~\cite{google2026gemini4argon}, run only under \texttt{bare} and \texttt{axes} owing to API cost; and
(iii)~\emph{open-weight VLMs}: Qwen3-VL~\cite{qwen3-vl}, Gemma 4~\cite{gemma4}, and InternVL3.5~\cite{internvl35}, covering dense and MoE architectures from 1B to 38B parameters.

\noindent
\textbf{Metric.}
In the synthetic dataset, each episode has two positive and six negative queries, so we report Balanced Accuracy (BA)~\cite{ba}, with 95\% confidence intervals from bootstrap resampling at the level of $\mathcal{V}(I^+)$ to account for episodes sharing the same positive-example attribute combination.

\begin{table}[t]
\centering
\caption{\textbf{Main results (Balanced Accuracy).}
Base VLMs use \texttt{axes}; vision encoders use no prompt.
95\% bootstrap confidence intervals are shown in brackets.
Bold denotes the highest reported score in each performance column.}
\label{tab:main_results}

\footnotesize
\setlength{\tabcolsep}{2pt}
\newcommand{\baci}[2]{#1\,{\scriptsize[#2]}}
\begin{tabular*}{\linewidth}{@{\extracolsep{\fill}}l c ccc@{}}
\toprule
& \textbf{Overall}
& \multicolumn{3}{c}{\textbf{Target condition}} \\
\cmidrule(lr){2-2}\cmidrule(lr){3-5}
\textbf{Model}
& \textbf{BA}
& \textbf{Shape} & \textbf{Texture} & \textbf{Pattern} \\
\midrule
Random Chance
& 0.500 & 0.500 & 0.500 & 0.500 \\
\midrule
\rowcolor{headergray}
\multicolumn{5}{@{}l}{\textit{Vision encoders (no prompt)}} \\
CLIP ViT-L/14~\cite{clip}
& \baci{0.608}{0.60, 0.62} & 0.721 & 0.561 & 0.541 \\
DINOv2-L~\cite{dinov2}
& \baci{0.608}{0.60, 0.62} & 0.746 & 0.530 & 0.547 \\
SigLIP so400m~\cite{siglip}
& \baci{0.615}{0.61, 0.62} & 0.703 & 0.603 & 0.539 \\
\midrule
\rowcolor{headergray}
\multicolumn{5}{@{}l}{\textit{Commercial VLMs}} \\
GPT-6.1 Terra~\cite{openai2026gpt61terra}
& \baci{0.886}{0.88, 0.90} & 0.938 & 0.907 & 0.812 \\
Gemini 4 Argon~\cite{google2026gemini4argon}
& \baci{\textbf{0.904}}{0.89, 0.91}
& \textbf{0.962} & \textbf{0.929} & \textbf{0.820} \\
\midrule
\rowcolor{headergray}
\multicolumn{5}{@{}l}{\textit{Open-weight VLMs}} \\
\rowcolor{accentpurple}
\multicolumn{5}{@{}l}{\textit{Qwen3-VL~\cite{qwen3-vl}}} \\
2B
& \baci{0.530}{0.52, 0.54} & 0.538 & 0.554 & 0.497 \\
4B
& \baci{0.518}{0.51, 0.52} & 0.440 & 0.607 & 0.506 \\
8B
& \baci{0.563}{0.55, 0.57} & 0.554 & 0.621 & 0.514 \\
30B-A3B
& \baci{0.515}{0.51, 0.52} & 0.477 & 0.543 & 0.525 \\
32B
& \baci{0.659}{0.65, 0.67} & 0.782 & 0.684 & 0.509 \\
\hdashline
\rowcolor{accentpurple}
\multicolumn{5}{@{}l}{\textit{Gemma 4~\cite{gemma4}}} \\
E2B
& \baci{0.522}{0.51, 0.53} & 0.586 & 0.492 & 0.489 \\
E4B
& \baci{0.505}{0.50, 0.51} & 0.550 & 0.489 & 0.476 \\
12B
& \baci{0.610}{0.60, 0.62} & 0.636 & 0.638 & 0.555 \\
31B
& \baci{0.671}{0.66, 0.68} & 0.762 & 0.665 & 0.586 \\
\hdashline
\rowcolor{accentpurple}
\multicolumn{5}{@{}l}{\textit{InternVL3.5~\cite{internvl35}}} \\
1B
& \baci{0.550}{0.54, 0.56} & 0.656 & 0.507 & 0.487 \\
2B
& \baci{0.529}{0.52, 0.54} & 0.527 & 0.535 & 0.525 \\
4B
& \baci{0.512}{0.50, 0.52} & 0.530 & 0.509 & 0.496 \\
8B
& \baci{0.522}{0.52, 0.53} & 0.515 & 0.534 & 0.518 \\
14B
& \baci{0.511}{0.51, 0.51} & 0.515 & 0.515 & 0.504 \\
30B-A3B
& \baci{0.527}{0.52, 0.53} & 0.489 & 0.535 & 0.556 \\
38B
& \baci{0.617}{0.61, 0.62} & 0.719 & 0.560 & 0.571 \\
\bottomrule
\end{tabular*}
\par\smallskip
\begin{minipage}{\linewidth}
\end{minipage}
\end{table}

\subsection{Results on the Synthetic Gavagai Benchmark}

Table~\ref{tab:main_results} shows the overall balanced accuracy across the evaluated models.

\noindent
\textbf{Overall Comparisons.}
Commercial VLMs substantially outperform open VLMs, with the best commercial models approaching 0.9 balanced accuracy.
Among open models, larger VLMs generally perform better, whereas several smaller VLMs even underperform vision encoder baselines.

This suggests that strong reasoning capability, rather than simply adopting a VLM architecture, is important for reasoning from negative evidence.



\noindent
\textbf{Performance Across Visual Attributes.}
Performance varies substantially across visual attributes.
Vision encoder baselines strongly favor shape over texture and pattern.
A similar tendency is observed in the larger open-weight VLMs, where shape generally remains the strongest attribute.
Commercial VLMs improve performance across all attributes, while the same overall ordering remains.

These results suggest that stronger reasoning can reduce the performance gap across attributes, while texture and pattern remain more challenging than shape, indicating that part of the imbalance may also stem from differences in attribute difficulty.


\begin{table}[t]
\centering
\caption{\textbf{Effect of prior information and reasoning strategies (Balanced Accuracy).}
Overall scores across four prior settings (\texttt{bare}, \texttt{rule}, \texttt{axes}, \texttt{values}) and two reasoning strategies (\texttt{CoT}, \texttt{HEA}).
Bold denotes the highest score within each model row, including ties.}
\label{tab:instruction_results}

\scriptsize
\setlength{\tabcolsep}{2pt}
\begin{tabular*}{\linewidth}{@{\extracolsep{\fill}}l cccc cc@{}}
\toprule
& \multicolumn{4}{c}{\textbf{Prior}} & \multicolumn{2}{c}{\textbf{Reasoning}} \\
\cmidrule(lr){2-5} \cmidrule(lr){6-7}
\textbf{Model}
& \texttt{bare}
& \texttt{rule}
& \texttt{axes}
& \texttt{values}
& \texttt{CoT}
& \texttt{HEA} \\
\midrule
\rowcolor{headergray}
\multicolumn{7}{@{}l}{\textit{Commercial VLMs}} \\
GPT-6.1 Terra~\cite{openai2026gpt61terra}
& 0.714 & --- & \textbf{0.886} & --- & --- & --- \\
Gemini 4 Argon~\cite{google2026gemini4argon}
& 0.748 & --- & \textbf{0.904} & --- & --- & --- \\
\midrule
\rowcolor{headergray}
\multicolumn{7}{@{}l}{\textit{Open-weight VLMs}} \\
Qwen3-VL-8B~\cite{qwen3-vl}
& 0.570 & 0.569 & 0.563 & 0.569 & 0.540 & \textbf{0.606} \\
Qwen3-VL-32B~\cite{qwen3-vl}
& 0.551 & 0.590 & 0.659 & 0.652 & 0.595 & \textbf{0.693} \\
\hdashline
Gemma 4 E4B~\cite{gemma4} & 0.520 & 0.518 & 0.505 & 0.438 & 0.536 & \textbf{0.601} \\
Gemma 4 31B~\cite{gemma4}
& 0.605 & 0.631 & 0.671 & \textbf{0.677} & 0.625 & 0.658 \\
\hdashline
InternVL3.5-8B~\cite{internvl35} & 0.535 & 0.543 & 0.522 & 0.518 & 0.521 & \textbf{0.601} \\
InternVL3.5-38B~\cite{internvl35}
& 0.584 & 0.607 & 0.617 & 0.584 & 0.552 & \textbf{0.660} \\
\bottomrule
\end{tabular*}
\end{table}

\noindent
\textbf{Effect of Task Priors.}
Table~\ref{tab:instruction_results} shows that even strong commercial VLMs exhibit a substantial gap between \texttt{bare} and \texttt{axes}, suggesting that current models struggle to construct the appropriate hypothesis space without task-specific prior.
This effect is capability-dependent: smaller VLMs benefit little from structural priors, whereas larger models exploit them more effectively, consistent with prior findings that task-specific cues can elicit capabilities that are only partially expressed under underspecified instructions~\cite{xu2026decoycount,van2026hinted}.

Interestingly, providing the full set of candidate values does not consistently improve over \texttt{axes}, suggesting that the benefit comes primarily from exposing the relevant structure of the hypothesis space rather than simply providing more information.
Together, these results highlight the importance of identifying the appropriate visual hypothesis space for successful novel concept recognition.

\noindent
\textbf{Effect of Reasoning Guidance.}
Table~\ref{tab:instruction_results} further investigates whether explicit reasoning guidance facilitates novel concept recognition.
Although \texttt{CoT} instructs models to generate and eliminate hypotheses within a single reasoning pass, it yields inconsistent improvements over \texttt{bare}, and even degrades performance for some models.
In contrast, \texttt{HEA}, which explicitly separates hypothesis generation, elimination, and application into distinct model calls, consistently outperforms \texttt{CoT} across the three largest open-weight VLMs, with gains of up to 10.8 percentage points.

Notably, \texttt{HEA} also outperforms \texttt{axes} on two of the three models, despite receiving no predefined attribute categories or values.
These findings suggest that explicitly separating hypothesis generation and evidence-based elimination into distinct stages can facilitate novel concept recognition, even without task-specific structural priors.
More broadly, the results highlight the potential importance of intermediate hypothesis verification for integrating visual evidence with language-based reasoning.

\begin{table}[t]
\centering
\caption{\textbf{Performance by negative category (Balanced Accuracy).}
Base VLMs use \texttt{axes}; vision encoders use no prompt.
$\Delta$ is F minus P, computed from the displayed scores.
Bold denotes the highest reported score in each of the F and P columns.}
\label{tab:negative_results}

\footnotesize
\setlength{\tabcolsep}{2pt}
\begin{tabular*}{\linewidth}{@{\extracolsep{\fill}}l rrr@{}}
\toprule
\textbf{Model} & \textbf{F} & \textbf{P} & $\Delta$ \\
\midrule
\rowcolor{headergray}
\multicolumn{4}{@{}l}{\textit{Vision encoders (no prompt)}} \\
CLIP ViT-L/14~\cite{clip}
& 0.633 & 0.577 & $+0.056$ \\
DINOv2-L~\cite{dinov2}
& 0.638 & 0.570 & $+0.068$ \\
SigLIP so400m~\cite{siglip}
& 0.642 & 0.581 & $+0.061$ \\
\midrule
\rowcolor{headergray}
\multicolumn{4}{@{}l}{\textit{Commercial VLMs}} \\
GPT-6.1 Terra~\cite{openai2026gpt61terra}
& 0.896 & 0.874 & $+0.022$ \\
Gemini 4 Argon~\cite{google2026gemini4argon}
& \textbf{0.906} & \textbf{0.902} & $+0.004$ \\
\midrule
\rowcolor{headergray}
\multicolumn{4}{@{}l}{\textit{Open-weight VLMs}} \\
Qwen3-VL-8B~\cite{qwen3-vl}
& 0.569 & 0.555 & $+0.014$ \\
Qwen3-VL-32B~\cite{qwen3-vl}
& 0.683 & 0.628 & $+0.055$ \\
\quad +HEA (ours) & 0.703 & 0.680 & $+0.023$ \\
\hdashline
Gemma 4 E4B~\cite{gemma4} & 0.527 & 0.477 & $+0.050$ \\
Gemma 4 31B~\cite{gemma4}
& 0.711 & 0.621 & $+0.090$ \\
\quad +HEA (ours) & 0.687 & 0.623 & $+0.064$ \\
\hdashline
InternVL3.5-8B~\cite{internvl35} & 0.523 & 0.521 & $+0.002$ \\
InternVL3.5-38B~\cite{internvl35}
& 0.629 & 0.601 & $+0.028$ \\
\quad +HEA (ours)
& 0.665 & 0.654 & $+0.011$ \\
\bottomrule
\end{tabular*}
\par\smallskip
\begin{minipage}{\linewidth}
\footnotesize
\end{minipage}
\end{table}

\begin{table}[t]
\centering
\caption{
\textbf{Decomposition of capability gaps} on the synthetic benchmark (BA).
Oracle Attributes (\textbf{OA}) provides ground-truth visual attributes for all images, Predicted Attributes (\textbf{PA}) uses the model's own attribute predictions, and Oracle Rule (\textbf{OR}) provides the ground-truth decision rule.
$\Delta_{\mathrm{reason}}$ and $\Delta_{\mathrm{perc}}$ quantify the reasoning and perception gaps, respectively.
The larger gap is highlighted in bold. Gaps are computed before rounding.
}
\label{tab:cap-gaps}

\footnotesize
\setlength{\tabcolsep}{2pt}
\begin{tabular*}{\linewidth}{
@{\extracolsep{\fill}}lcccccc@{}
}
\toprule
\textbf{Model}
& \texttt{axes}
& OR
& PA
& OA
& $\Delta_{\mathrm{reason}}$
& $\Delta_{\mathrm{perc}}$ \\
\midrule
\rowcolor{accentpurple}
\multicolumn{7}{@{}l}{\textit{Qwen3-VL}~\cite{qwen3-vl}} \\
\quad 2B      & 0.530 & 0.976 & 0.450 & 0.460
        & \textbf{0.516} & 0.011 \\
\quad 4B      & 0.518 & 1.000 & 0.587 & 0.683
        & \textbf{0.317} & 0.097 \\
\quad 8B      & 0.563 & 1.000 & 0.572 & 0.610
        & \textbf{0.390} & 0.038 \\
\quad 30B-A3B & 0.515 & 1.000 & 0.559 & 0.555
        & \textbf{0.445} & -0.004 \\
\quad 32B     & 0.659 & 1.000 & 0.714 & 0.777
        & \textbf{0.223} & 0.063 \\
\hdashline
\rowcolor{accentpurple}
\multicolumn{7}{@{}l}{\textit{Gemma 4}~\cite{gemma4}} \\
\quad E2B     & 0.522 & 1.000 & 0.520 & 0.533
        & \textbf{0.467} & 0.014 \\
\quad E4B     & 0.505 & 1.000 & 0.514 & 0.578
        & \textbf{0.422} & 0.065 \\
\quad 12B     & 0.610 & 1.000 & 0.541 & 0.553
        & \textbf{0.447} & 0.012 \\
\quad 31B     & 0.671 & 1.000 & 0.719 & 0.849
        & \textbf{0.151} & 0.129 \\
\hdashline
\rowcolor{accentpurple}
\multicolumn{7}{@{}l}{\textit{InternVL3.5}~\cite{internvl35}} \\
\quad 1B      & 0.550 & 0.675 & 0.528 & 0.576
        & \textbf{0.099} & 0.048 \\
\quad 2B      & 0.529 & 0.981 & 0.517 & 0.675
        & \textbf{0.306} & 0.158 \\
\quad 4B      & 0.512 & 1.000 & 0.537 & 0.691
        & \textbf{0.309} & 0.154 \\
\quad 8B      & 0.522 & 1.000 & 0.539 & 0.824
        & 0.176 & \textbf{0.285} \\
\quad 14B     & 0.511 & 1.000 & 0.563 & 0.906
        & 0.094 & \textbf{0.343} \\
\quad 30B-A3B & 0.527 & 1.000 & 0.563 & 0.782
        & 0.218 & \textbf{0.220} \\
\quad 38B     & 0.617 & 1.000 & 0.681 & 0.959
        & 0.041 & \textbf{0.278} \\
\bottomrule
\end{tabular*}
\end{table}

\noindent
\textbf{Effect of Negative Evidence.}
We compare full-elimination (F) and partial-elimination (P) to examine how the structure of negative evidence affects reasoning. An F-negative differs from the positive example only in the target attribute, allowing competing hypotheses to be eliminated from a single example. In contrast, each P-negative eliminates only one candidate attribute, requiring evidence to be integrated across multiple negatives. Table~\ref{tab:negative_results} reports average performance, while Fig.~\ref{fig:scaling} in the supplementary material shows how performance changes with the number of negatives \(J\).

F consistently outperforms P for most models, with particularly clear gaps for stronger open-weight VLMs. As shown in Fig.~\ref{fig:scaling}, F yields a large improvement with only one negative and then tends to saturate, whereas P improves more gradually as additional negatives are provided. The improvement under P indicates that models can integrate evidence across multiple examples to some extent. However, the gains remain limited and generally do not close the gap to F, suggesting that cross-example integration is still challenging. Commercial VLMs show a much smaller F--P gap, indicating stronger ability to aggregate negative evidence across examples.

\noindent
\textbf{Decomposition of Perception and Reasoning.}
To identify whether model errors arise from perception or negative-based reasoning, we consider three diagnostic conditions: Oracle Attributes (\textbf{OA}) provides ground-truth visual attributes for all images, Predicted Attributes (\textbf{PA}) uses the model's own attribute predictions, and Oracle Rule (\textbf{OR}) provides the ground-truth rule. We quantify the reasoning gap $\Delta_{\mathrm{reason}}$ as the performance difference between Oracle Rule and Oracle Attributes, and the perception gap $\Delta_{\mathrm{perc}}$ as that between Oracle Attributes and Predicted Attributes (see Sec.~\ref{sec:supp-bottleneck} of the supplementary material for details).

As shown in Table~\ref{tab:cap-gaps}, Qwen3-VL and Gemma 4 consistently exhibit larger reasoning gaps than perception gaps, indicating that eliminating inconsistent hypotheses remains a major challenge even when visual attributes are explicitly provided. In contrast, InternVL3.5 exhibits a shift in its primary bottleneck with model scale: reasoning gaps are larger for smaller models, whereas perception gaps dominate for models with 8B parameters or more. Notably, InternVL3.5 38B has a reasoning gap of only 0.041, compared with a perception gap of 0.278. These results suggest that the relative contributions of perception and reasoning vary across model families and scales, with negative-based reasoning remaining a major limitation for Qwen3-VL and Gemma 4, while visual perception becomes the primary bottleneck for larger InternVL3.5 models.

\begin{table}[t]
\centering
\caption{\textbf{Balanced Accuracy ($\mathrm{BA}_r$) across negative taxonomic ranks on newly described species.} Episodes are evaluated under the \texttt{bare} instruction ($J=3$). Bold denotes the highest score in each column, including ties.}
\label{tab:natural}
\footnotesize
\setlength{\tabcolsep}{4.2pt}
\begin{tabular}{@{}lcccccc@{}}
\toprule
\textbf{Model} & \textbf{Genus} & \textbf{Family} & \textbf{Order} & \textbf{Class} & \textbf{Phylum} & \textbf{Avg.} \\
\midrule
\rowcolor{headergray}
\multicolumn{7}{@{}l}{\textit{Vision Encoder Baselines (No Prompt)}} \\
\quad CLIP~\cite{clip} & 0.654 & 0.652 & 0.686 & 0.875 & 0.740 & 0.722 \\
\quad DINOv2~\cite{dinov2}    & 0.639 & 0.677 & 0.723 & 0.918 & 0.818 & 0.755 \\
\quad SigLIP~\cite{siglip} & 0.684 & 0.683 & 0.697 & 0.902 & 0.717 & 0.736 \\
\quad BioCLIP 2~\cite{gu2025bioclip2} & 0.682 & 0.730 & 0.763 & 0.920 & 0.826 & 0.784 \\
\midrule
\rowcolor{headergray}
\multicolumn{7}{@{}l}{\textit{Commercial VLMs}} \\
\quad GPT-6.1 Terra  & 0.757 & 0.791 & 0.829 & 0.955 & 0.975 & 0.862 \\
\quad Gemini 4 Argon & \textbf{0.762} & \textbf{0.804} & 0.822 & 0.944 & 0.926 & 0.852 \\
\midrule
\rowcolor{headergray}
\multicolumn{7}{@{}l}{\textit{Open-weight VLMs}} \\
\rowcolor{accentpurple}
\multicolumn{7}{@{}l}{\textit{Qwen3-VL}~\cite{qwen3-vl}} \\
\quad 2B      & 0.498 & 0.496 & 0.525 & 0.512 & 0.500 & 0.506 \\
\quad 4B      & 0.650 & 0.636 & 0.681 & 0.822 & 0.682 & 0.694 \\
\quad 8B      & 0.699 & 0.735 & 0.780 & 0.929 & 0.896 & 0.808 \\
\quad 30B-A3B & 0.622 & 0.655 & 0.731 & 0.900 & 0.872 & 0.756 \\
\quad 32B     & 0.722 & 0.788 & 0.827 & \textbf{0.971} & 0.961 & 0.854 \\
\quad +HEA (ours) & 0.723 & 0.789 & 0.850 & \textbf{0.971} & \textbf{0.992} & \textbf{0.865} \\
\hdashline
\rowcolor{accentpurple}
\multicolumn{7}{@{}l}{\textit{Gemma 4}~\cite{gemma4}} \\
\quad E2B & 0.595 & 0.624 & 0.662 & 0.744 & 0.849 & 0.695 \\
\quad E4B & 0.580 & 0.600 & 0.668 & 0.812 & 0.726 & 0.677 \\
\quad 12B & 0.734 & 0.791 & 0.846 & 0.964 & 0.971 & 0.861 \\
\quad 31B & 0.723 & 0.772 & 0.831 & 0.950 & 0.893 & 0.834 \\
\quad +HEA (ours) & 0.720 & 0.777 & \textbf{0.853} & 0.955 & 0.968 & 0.855 \\
\hdashline
\rowcolor{accentpurple}
\multicolumn{7}{@{}l}{\textit{InternVL3.5}~\cite{internvl35}} \\
\quad 1B      & 0.572 & 0.567 & 0.607 & 0.777 & 0.809 & 0.667 \\
\quad 2B      & 0.543 & 0.529 & 0.565 & 0.723 & 0.711 & 0.614 \\
\quad 4B      & 0.687 & 0.683 & 0.713 & 0.832 & 0.773 & 0.738 \\
\quad 8B      & 0.700 & 0.733 & 0.773 & 0.917 & 0.947 & 0.814 \\
\quad 14B     & 0.688 & 0.744 & 0.779 & 0.924 & 0.950 & 0.817 \\
\quad 30B-A3B & 0.622 & 0.619 & 0.680 & 0.806 & 0.764 & 0.698 \\
\quad 38B     & 0.734 & 0.767 & 0.804 & 0.954 & 0.974 & 0.847 \\
\quad +HEA (ours) & 0.654 & 0.735 & 0.814 & 0.956 & 0.985 & 0.829 \\
\bottomrule
\end{tabular}
\end{table}

\subsection{Results on the Real Gavagai Benchmark}
We next examine whether the findings from the Synthetic Gavagai Benchmark also hold for real-world visual concepts. Unlike the synthetic setting, the Real Gavagai Benchmark does not guarantee support-set sufficiency because visual attributes cannot be explicitly controlled. It instead evaluates concept recognition under partial negative evidence. We additionally include BioCLIP 2~\cite{gu2025bioclip2}, a domain-specific vision encoder trained on organism images with hierarchical taxonomic supervision.


\noindent
\textbf{Overall Performance.}
Table~\ref{tab:natural} reports the results on the Real
Gavagai Benchmark.
As in the synthetic setting, strong VLMs generally
outperform smaller VLMs and vision encoder baselines.
Among the vision encoders, BioCLIP 2 achieves the highest
average balanced accuracy of 0.784, outperforming the
general-purpose encoders.
This improvement suggests that visual representations
learned from domain-specific biological data are useful for
identifying novel species.

However, the strongest VLMs achieve average scores above
0.85, outperforming even BioCLIP 2.
This result suggests that, in addition to visual
representations suited to the domain, reasoning over
positive and negative examples to identify discriminative
attributes is important for recognizing novel species.


\noindent
\textbf{Reasoning Guidance and Its Limitations.}
Explicit reasoning with HEA improves the average balanced
accuracy of Qwen3-VL 32B from 0.854 to 0.865 and Gemma 4
31B from 0.834 to 0.855.
These gains suggest that models can better recognize novel concepts in real images by explicitly identifying candidate attributes and eliminating those inconsistent with negative examples.

However, HEA provides little or no improvement at the
genus level, where distinguishing closely related species
may require perceiving subtle visual differences.
Moreover, HEA does not improve InternVL3.5 38B, for which visual perception remains the dominant bottleneck on the synthetic dataset (Table~\ref{tab:cap-gaps}).
These results suggest that reasoning alone may be insufficient when perception is the primary bottleneck.

\begin{figure}
    \centering
    \includegraphics[width=\linewidth]{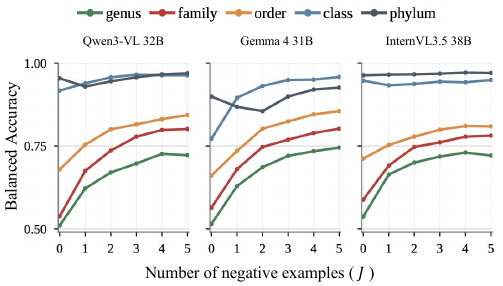}
    \caption{\textbf{Effect of the number of negative examples.}
    Increasing $J$ generally improves performance, particularly for genus, family, and order level negatives.}
    \label{fig:negative_curve}
\end{figure}

\noindent
\textbf{Reasoning from Negative Evidence in Real Images.}
We further investigate whether VLMs can leverage negative evidence
to resolve concept ambiguity in natural images.
Figure~\ref{fig:negative_curve} shows the effect of varying
the number of negative examples ($J$) across different taxonomic ranks.
Across all three VLMs, balanced accuracy generally increases
as more negative examples are provided, with particularly
large improvements at finer taxonomic ranks.
This trend is consistent with our formulation of negative-based
reasoning: each additional negative example provides evidence
for ruling out candidate hypotheses, progressively narrowing the space
of plausible interpretations.
These findings demonstrate that negative-based reasoning remains effective for novel concept recognition, even when the defining visual attributes must be inferred from complex natural images.

\begin{figure}
    \centering
    \includegraphics[width=\linewidth]{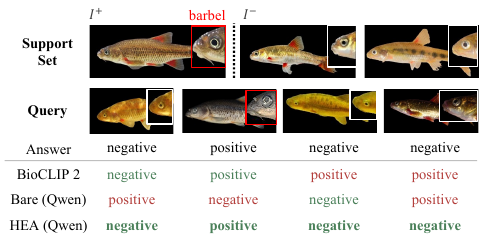}
    \caption{\textbf{Example of visual reasoning for novel concept classification.} The VLM identifies discriminative visual regions by comparing positive and negative support examples and uses these regions to classify query images as positive or negative instances of the target concept.}
    \label{fig:qualitative}
\end{figure}

\noindent
\textbf{Qualitative Analysis.}
Figure~\ref{fig:qualitative} illustrates hypothesis elimination on the Real Gavagai Benchmark. Our method identifies a barbel near the mouth as a discriminative attribute absent from negative examples and correctly classifies all queries based on this feature. In contrast, BioCLIP 2 and the bare VLM struggle to distinguish this subtle feature. This result highlights the importance of negative-based reasoning beyond visual similarity.

\section{Limitations}
Our benchmark has three main limitations. First, as shown in our genus-level analysis, recognizing novel species requires not only hypothesis elimination but also hypothesis generation through visual perception. Second, our benchmark evaluates independent episodes without considering how eliminated hypotheses and acquired knowledge can be retained and reused for subsequent learning. Third, our benchmark focuses on animal-related concepts, leaving extensions to other domains, such as medical imaging and robotics, for future work.

\section{Conclusion}
In this paper, we introduced the Gavagai Benchmark to evaluate novel visual concept recognition through negative-based reasoning. By constructing synthetic and real-world datasets, we systematically investigated how VLMs identify defining visual attributes from one positive and a few negative examples. Our experiments revealed that negative-based reasoning remains a major bottleneck for several VLM families, particularly when integrating evidence across multiple negatives, while visual perception also poses challenges in real-world settings. These findings highlight the importance of both accurate perception and effective negative-based reasoning for recognizing novel visual concepts.


\section*{LLM Usage Statement}
We used LLM tools to polish text, assist with parts of the experimental and
analysis code, and help produce figures from our results. All LLM-assisted work was reviewed by the authors, who take full responsibility for the final text, claims, code, and figures.

{
    \small
    \bibliographystyle{ieeenat_fullname}
    \bibliography{main}
}
\clearpage
\setcounter{page}{1}
\maketitlesupplementary

\appendix

\section{Comparison with Existing Datasets}
\label{sec:supp-hypothesis-space}

Quine's gavagai problem illustrates the ambiguity of learning a concept from an observation: when a word is uttered in the presence of a rabbit, it may refer to the rabbit, a part of it, or another property of the scene~\cite{Quine1960}. In our visual formulation, one positive image is similarly compatible with multiple sets of attributes. Observing that an image belongs to a concept does not reveal which of its attributes define that concept.

As defined in Sec.~\ref{sec:problem_formulation}, every nonempty subset of $\mathcal{V}(I^+)$ is a candidate hypothesis. A negative image eliminates a candidate only when it contains \emph{all} attributes in that candidate. Thus, an attribute shared with a negative cannot define the concept on its own, but may still appear in a consistent multi-attribute hypothesis. Several hypotheses may remain even after observing the negatives; the support set is sufficient for a query set if those hypotheses agree on its labels, not necessarily if a unique hypothesis remains.

\begin{figure}[t]
    \centering
    \includegraphics[width=\linewidth]{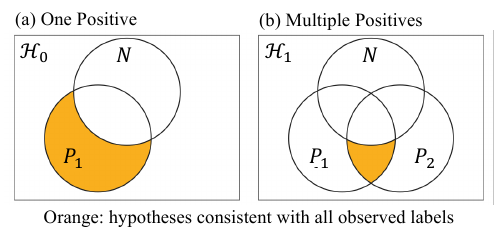}
    \caption{\textbf{Effect of an additional positive example on the hypothesis space.} Each region contains candidate concepts, where each candidate is a set of attributes. $P_i$ contains the hypotheses compatible with positive image $I_i^+$; $N$ contains hypotheses incompatible with the negative image. Orange denotes the hypotheses consistent with all observed labels: $P_1\setminus N$ with one positive and $(P_1\cap P_2)\setminus N$ with two. The diagrams are schematic; an additional positive does not always eliminate candidates.}
    \label{fig:supp-hypothesis-space}
\end{figure}

\noindent
\textbf{One versus multiple positives.}
Let $\mathcal{H}_0$ be the universe of candidate concepts, and let $P_i=\{h\in\mathcal{H}_0\mid h\subseteq\mathcal{V}(I_i^+)\}$ contain hypotheses consistent with positive image $I_i^+$. For a negative image $I^-$, let $N=\{h\in\mathcal{H}_0\mid h\subseteq\mathcal{V}(I^-)\}$ contain hypotheses it rules out. Figure~\ref{fig:supp-hypothesis-space} shows that the consistent hypotheses are $P_1\setminus N$ with one positive and $(P_1\cap P_2)\setminus N$ with two positives.

For positive examples $I_1^+,\ldots,I_K^+$, a hypothesis must be contained in every positive image. Consequently, the initial space is the set of nonempty subsets of $\bigcap_{i=1}^{K}\mathcal{V}(I_i^+)$. Adding a positive can only retain or remove candidates; it does not necessarily remove any. With one positive, no such intersection across positives is available, so the negative examples provide the evidence for rejecting candidates that fit that positive.

For example, let the first positive contain attributes $\{a,b,c\}$ and the negative contain $\{b\}$. Of the seven nonempty subsets of the positive, the negative eliminates only $\{b\}$; in particular, $\{a,b\}$ remains possible. If a second positive contains $\{a,b\}$, candidates containing $c$ are also ruled out, leaving $\{a\}$ and $\{a,b\}$. The additional positive narrows the candidate space even though the negative is unchanged.

This comparison clarifies why our single-positive setting emphasizes negative evidence. It does not imply that multiple-positive settings such as Bongard problems can be solved without negatives: shared positive attributes can still be ambiguous, and negative examples can reject inconsistent rules. In the synthetic dataset we additionally restrict candidates to single attributes, allowing the support negatives to isolate one controlled attribute; the general formulation above also permits conjunctions.

Few-shot visual classification evaluates generalization from labeled support images, while Bongard-style benchmarks require a visual rule to be inferred from positive and negative sets~\cite{prototype,bongardopenworld}. Our synthetic dataset isolates the ambiguity of a single positive image: the same positive and queries are paired with different negative sets, changing the attribute that defines the target concept. Its F- and P-negative configurations additionally vary whether one negative or a combination of negatives is sufficient to eliminate the non-target attributes. The real dataset extends the evaluation to recently described species and varies the taxonomic relatedness of negatives. Unlike the synthetic conditions, the target species remains fixed across ranks.

\noindent
\textbf{Novel concepts.}
The concepts in Bongard-style benchmarks are known to the model and can be named: each problem has a solution that can be stated in words, such as a geometric property in the original Bongard problems~\cite{bongard,bongardlogo}, a human--object interaction in Bongard-HOI~\cite{bongardhoi}, or a free-form phrase in Bongard-OpenWorld~\cite{bongardopenworld}.
Such a problem can in principle be solved by describing the images in text and finding the phrase that separates the two sets.
The Real Gavagai Benchmark instead targets species formally described since 2025, which are unlikely to appear under their names in the training data of the models, so the concept is novel to the model.
The attributes that distinguish a new species from its relatives often have no established name, and the rule may therefore not be expressible in text at all.
Moreover, the model may not perceive the distinguishing attributes in the first place, since its visual representation was never trained to separate the new species from its relatives.
Recognizing a novel concept thus requires using the visual evidence in the support set directly, beyond what the model can verbalize.
The Synthetic Gavagai Dataset complements this with known, nameable attributes, so that the ambiguity of a single positive can be studied under control.

\noindent
\textbf{A realistic setting.}
Beyond its difficulty, our benchmark differs from existing ones in the setting it assumes.
Bongard-style problems are puzzles: several positives and negatives are curated so that one rule separates them.
We instead model an agent that lives in the real world and encounters something it does not know.
Such an agent typically sees a single example of the new concept, e.g., when someone points to an organism and names it, and can compare it with things it already knows to be different, such as related species; it must then recognize new individuals of the concept under natural variation in pose, appearance, and viewpoint.
Our formulation follows this situation: one positive example, a few negatives from related categories at varying taxonomic distance, and queries that are other individuals of the novel species or of its relatives.
The benchmark thus evaluates the capability such an agent needs, namely learning a novel visual concept from a single encounter by resolving its ambiguity with negative evidence.

\section{Synthetic Gavagai Dataset Detail}
\label{sec:supp-synthetic-dataset-detail}

\subsection{Visual Attribute Set}
Table~\ref{tab:concepts} lists the candidate values for each controlled visual factor. Shape describes the animal's body form, texture its surface material, and pattern its surface markings. The factors are combined to create images with distinct attribute combinations.

\begin{table}[tb]
\caption{Visual attribute sets.}
\label{tab:concepts}
\hbox to\hsize{\hfil
\begin{tabular}{lp{0.70\hsize}}\toprule
Visual attribute & Values \\\midrule
Shape (8) & Cat, dog, horse, rabbit, squirrel, macaque, pig, dromedary camel \\
Texture (5) & Elephant skin, sheep wool, snake scales, short fur, bird feathers \\
Pattern (5) & Plain, giraffe spots, tiger stripes, leopard spots, cow patches \\\bottomrule
\end{tabular}\hfil}
\end{table}

\subsection{Image Generation}
\label{sec:image-generation}

We synthesize images using an image generation model (gemini-3.1-flash-image) through the following three-stage pipeline.

\noindent
\textbf{Base images}: 
For each combination of shape and imaging conditions, we generate an image of an animal with its original body surface, resulting in 64 base images.

\noindent
\textbf{Reference images}: 
For each combination of texture and color, we first generate a close-up image of an animal surface without any pattern. We then edit the image to add a pattern, producing 100 close-up images corresponding to combinations of texture, color, and pattern. 

\noindent
\textbf{Composite images}: 
We replace only the body surface of each base image using a randomly cropped region from a reference image. The editing instructions explicitly require the original shape, pose, and background to be preserved, without mentioning the names of textures or patterns. This avoids introducing unintended shape attributes, such as ears or legs, associated with the animals from which the textures or patterns originate.

This process produces one synthetic image for each combination of shape, imaging conditions, texture, and pattern, yielding a total of 1,600 images.

\noindent
\textbf{Imaging conditions and colors}:
These are nuisance factors, varied so that no shape is tied to one pose or background and no texture to one color.
The eight imaging conditions per shape combine a background (\eg, grassy lawn, studio backdrop, snowy field, barn floor) with a camera view and posture (\eg, standing, walking, sitting, lying), assigned so that no background predicts a pose or a shape.
Each texture is rendered in four plausible colors (\eg, sheep wool in cream white, charcoal black, chestnut brown, or dove gray), so that no texture can be identified by its hue.

\noindent
\textbf{Human Verification}: 
Generated images may contain failures, such as missing patterns, residual regions of the original body surface, or distorted body shapes. In the initial verification round, four human annotators independently evaluated the generated images.

For each image, the annotators were shown the base image, the reference image used for surface replacement, the specified visual attributes, and the generated image. An image was accepted only if all three criteria were satisfied: (1) the shape, pose, and background were preserved from the base image; (2) the surface texture and pattern matched the reference image; and (3) the texture and pattern were applied naturally across the entire body.

Before the evaluation, the annotators completed a tutorial presenting three accepted and three rejected examples to clarify the evaluation criteria. Images accepted by all four annotators were retained. A subset of images was subsequently regenerated and verified again; the later verdict replaced the original verdict for these images. The resulting verified pool contains 1,103 images.

\subsection{Episode Construction}
For each positive image, we create three conditions, with shape, texture, or pattern as the target. The positive image and query set are shared across these conditions; only the support negatives and query labels change. Every image within an episode is distinct, and images can be reused across episodes.

Let $(s,t,p)$ denote the shape, texture, and pattern of the positive image. Tables~\ref{tab:supp-support-construction} and~\ref{tab:supp-query-construction} show the support and query constructions, respectively. Primes denote values different from the positive image's corresponding values; tildes denote query values that differ from both the positive and the negative-support values on the same axis.

\begin{table}[t]
\centering
\caption{\textbf{Synthetic support construction.} Entries are (shape, texture, pattern) tuples. The target condition determines which attribute is absent from each negative. $\mathrm{P}_1$ and $\mathrm{P}_2$ share different non-target attributes with the positive.}
\label{tab:supp-support-construction}
\small
\setlength{\tabcolsep}{3pt}
\begin{tabular}{@{}lccc@{}}
\toprule
& \textbf{Shape} & \textbf{Texture} & \textbf{Pattern} \\
\midrule
Positive & $(s,t,p)$ & $(s,t,p)$ & $(s,t,p)$ \\
F-negative & $(s',t,p)$ & $(s,t',p)$ & $(s,t,p')$ \\
$\mathrm{P}_1$ & $(s',t,p')$ & $(s,t',p')$ & $(s,t',p')$ \\
$\mathrm{P}_2$ & $(s',t',p)$ & $(s',t',p)$ & $(s',t,p')$ \\
\bottomrule
\end{tabular}
\end{table}

\noindent
\textbf{Support sufficiency.}
Under the single-attribute hypothesis prior, a support set is sufficient when its consistent hypotheses give the same label to every query. For example, in the shape condition, the candidate hypotheses are $\{s\}$, $\{t\}$, and $\{p\}$. The F-negative $(s',t,p)$ rules out both $\{t\}$ and $\{p\}$. Alternatively, $\mathrm{P}_1$ rules out $\{t\}$ and $\mathrm{P}_2$ rules out $\{p\}$. In either case, only $\{s\}$ remains, so all query labels are determined. The same argument applies to the other two target conditions. This conclusion relies on the single-attribute prior; the general formulation also permits conjunctions.

We construct F conditions with $J=1,\ldots,5$ negatives and P conditions with $J=2,\ldots,5$, including both P subtypes in each P condition. Each of the nine configurations contains 201 episodes, giving 1,809 episodes in total.

\begin{table}[t]
\centering
\caption{\textbf{Synthetic query construction.} Attribute tuples are ordered as (shape, texture, pattern). Each query type contains two images. Labels are determined by the target condition; the images are shared across the three conditions.}
\label{tab:supp-query-construction}
\small
\setlength{\tabcolsep}{4pt}
\begin{tabular}{@{}llccc@{}}
\toprule
& \textbf{Attributes} & \multicolumn{3}{c}{\textbf{Target condition}} \\
\cmidrule(lr){3-5}
\textbf{Type} & \textbf{Attributes} & \textbf{Shape} & \textbf{Texture} & \textbf{Pattern} \\
\midrule
A & $(s,\tilde{t},\tilde{p})$ & Pos. & Neg. & Neg. \\
B & $(\tilde{s},t,\tilde{p})$ & Neg. & Pos. & Neg. \\
C & $(\tilde{s},\tilde{t},p)$ & Neg. & Neg. & Pos. \\
D & $(\tilde{s},\tilde{t},\tilde{p})$ & Neg. & Neg. & Neg. \\
\bottomrule
\end{tabular}
\end{table}

\noindent
\textbf{Query construction.}
The ground-truth labels in Table~\ref{tab:supp-query-construction} are defined by the target attribute $a$, independently of support-set sufficiency: $I_q$ is positive if and only if $a\in\mathcal{V}(I_q)$. Types A, B, and C each share exactly one controlled attribute with $I^+$, and Type D shares none. Thus, $\{s\}$, $\{t\}$, and $\{p\}$ induce distinct labelings over $\mathcal{Q}$, whereas every multi-attribute conjunction in $\mathcal{H}^+\setminus\mathcal{H}_{\mathrm{single}}^+$ labels all queries negative. In this design, sufficiency under the single-attribute prior therefore also identifies the target attribute uniquely; allowing conjunctions leaves disagreement on the two target-positive queries.


\section{Real Gavagai Dataset Detail}
\label{sec:supp-real-dataset-detail}

\subsection{Dataset Collection}
\label{sec:real-gavagai-dataset-collection}
\noindent
\textbf{Data Collection.}
We collected 2,418 articles describing new species
published since 2025 from Novataxa.
Using a VLM-assisted filtering pipeline, we excluded
fossil taxa, cryptic species indistinguishable by
external morphology, and articles lacking natural
photographs.
For the remaining species, we cropped individual
organisms from figures in the original taxonomic
publications to construct positive examples and queries.

\begin{figure}[t]
    \centering
    \includegraphics[width=0.8\linewidth]{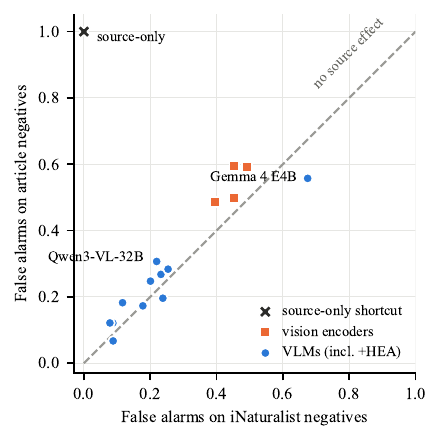}
    \caption{\textbf{Image-source shortcut check.} False-alarm rate on negative queries taken from articles vs.\ from iNaturalist (episodes with both kinds, averaged over ranks). A model answering by image source alone would lie at the top left; points on the diagonal show no source effect.}
    \label{fig:supp-source}
\end{figure}

To collect negative examples and queries, we retrieved
up to 10 research-grade observation images per
taxonomic rank from iNaturalist, selecting clear
depictions and cropping the organisms.
To reduce imaging-style shortcuts caused by using
different sources for positive and negative images,
we also incorporated images of other newly described
species from Novataxa when available.
Furthermore, we segmented all organisms using
SAM~3~\cite{sam3} and masked their backgrounds in black.
As shown in Fig.~\ref{fig:supp-source}, models do not
exploit the remaining source difference: false alarms on
article and iNaturalist negatives are close for every model,
far from a source-only classifier.
Because the article negatives come from taxonomic publications like the positives, they can share its imaging style, including preserved museum specimens; a model that accepted any specimen-like image would therefore show more false alarms on article negatives than on iNaturalist negatives, which we do not observe.

The final dataset comprises 297 newly described
species, including 285 animals and 12 fungi.
Each species has one positive example and two to four
positive queries.

\subsection{Episode Construction}
For each species and taxonomic rank, three negative support images are selected from the same group at that rank but a different group at the next finer rank. Positive queries are other images of the target species; negative queries depict other species in the same group at the selected rank. The numbers of positive and negative queries are equal, with two to four per class. The dataset contains 292 genus-level episodes and 297 at each of the four remaining ranks, for a total of 1,480. Negative support and query images are distinct, although their species can overlap.
Such an overlap could let a model reject a negative query by matching it to a conspecific negative in the support.
It is rare: 1.6\% of negative queries share their species with a negative support image (7.0\% at the genus level, 1.0\% at the family level, and none at coarser ranks).
Removing these queries changes the overall BA by at most 0.002 for Qwen3-VL 32B, Gemma 4 31B, InternVL3.5 38B, GPT-6.1 Terra, and Gemini 4 Argon.

\section{Structural Priors: Details}
\label{sec:supp-prompts}

\subsection{Prompts}
\label{sec:supp-prompt-text}

We evaluate each query independently. For all instruction
settings, the model receives a text instruction followed by
the labeled support images and one query image. Below,
bracketed text indicates an inserted image.

\paragraph{\texttt{bare}.}
\begin{quote}
Here are some example images, each labeled POSITIVE or
NEGATIVE.

Labeled examples:

Image 1 --- POSITIVE: [positive image]

Image 2 --- NEGATIVE: [negative image]

\ldots

Here is a new image.

Test image: [query image]

Is this new image POSITIVE or NEGATIVE? Answer with exactly
one word: ``positive'' or ``negative''.
\end{quote}

\paragraph{\texttt{rule}.}
The support images and query image are presented in the
same format as in \texttt{bare}. The initial instruction is
replaced with:

\begin{quote}
You are shown a set of labeled example images. Each is
labeled POSITIVE or NEGATIVE according to one single
hidden rule that defines a visual concept. Exactly one rule
explains every label you are shown. Work out that rule from
the examples, then apply it to classify new test images.
\end{quote}

The query is preceded by the following instruction:

\begin{quote}
Now classify this new test image using the SAME hidden
rule as above.
\end{quote}

After the query image, the model is instructed to answer:

\begin{quote}
Answer with exactly one word: ``positive'' or ``negative''.
\end{quote}

\paragraph{\texttt{axes}.}
This setting uses the same prompt as \texttt{rule}, with
the following instruction appended to its initial text:

\begin{quote}
Each image shows an animal that can be described by three
visual attributes: its SHAPE (which animal it is --- its
body, head and limbs), its TEXTURE (the material its body
surface is made of) and its PATTERN (the markings on that
surface). The hidden rule depends on exactly ONE of these
three attributes; the other two are irrelevant.
\end{quote}

\paragraph{\texttt{values}.}
This setting uses the same prompt as \texttt{axes}, with
the following instruction appended to its initial text:

\begin{quote}
The possible values are --- SHAPE: cat, dog, horse, rabbit,
squirrel, macaque monkey, pig, dromedary camel.
TEXTURE: wrinkled elephant hide, crimped sheep wool,
overlapping snake scales, short sleek fur, layered bird
feathers.
PATTERN: plain (no markings), giraffe-like net of patches,
tiger stripes, leopard rosettes, cow patches.
\end{quote}

\paragraph{\texttt{CoT}.}
This setting uses the same prompt as \texttt{bare}, so it names no rule, attribute, or value; only the final instruction is replaced with:

\begin{quote}
Is this new image POSITIVE or NEGATIVE? Think step by step before answering. First, describe each labeled example image in one short line. Second, describe the new image in one short line. Third, work out what distinguishes the POSITIVE examples from the NEGATIVE ones, and decide which group the new image belongs to. Finally, end your response with exactly one line: ``Answer: positive'' or ``Answer: negative''.
\end{quote}
The model may generate up to 1,024 tokens, and the label is read from the final line.

\subsection{Attribute Axes versus the Single-Attribute Constraint}
\label{sec:supp-prior-ablation}

The \texttt{axes} instruction gives the model two pieces of information at once: (A) the attribute axes along which the images vary (shape, texture, and pattern), and (B) the constraint that exactly one attribute defines the concept.
These are different interventions.
Without (B), the support set does not determine the labels of the queries: if conjunctions of attributes are allowed, a positive $\{s,t,p\}$ and a negative $\{s',t,p\}$ leave both $\{s\}$ and $\{s,t\}$ consistent, and the two disagree on a query that has $s$ but not $t$.
In the synthetic dataset this holds for every positive query, because the positive queries share only the target attribute with $I^+$ by construction; under conjunctive hypotheses, the labels of 25\% of all queries, namely all positive ones, are undetermined.
In the main paper (Effect of Task Priors in Sec.~\ref{sec:experiment}, Table~\ref{tab:instruction_results}), we interpret the gap between \texttt{bare} and \texttt{axes} as evidence that current models struggle to construct the appropriate hypothesis space themselves.
Part of this gap may instead come from resolving the underspecification, i.e., from (B) rather than from (A).

To separate the two, we evaluate two further instructions on all 1,809 episodes, both built on \texttt{rule}.
\texttt{single} adds only (B), without naming the axes:
\begin{quote}
The hidden rule depends on exactly ONE visual property of the animal; all its other properties are irrelevant.
\end{quote}
\texttt{axes\_conj} adds only (A), naming the axes as in \texttt{axes} but allowing combinations of them:
\begin{quote}
Each image shows an animal that can be described by three visual attributes: its SHAPE (\ldots), its TEXTURE (\ldots) and its PATTERN (\ldots). The hidden rule depends on one or more of these three attributes: on a single attribute, or on a combination of two or three of them.
\end{quote}
The descriptions of the axes are identical to those in \texttt{axes} (Sec.~\ref{sec:supp-prompt-text}).

Together with the instructions of the main paper (Sec.~\ref{sec:prompts}), the five instructions form the following comparisons, each of which isolates one piece of information:
\begin{description}
\item[\texttt{bare} $\rightarrow$ \texttt{rule}:] that one hidden rule explains all labels (the task framing).
\item[\texttt{rule} $\rightarrow$ \texttt{single}:] the single-attribute constraint (B) alone, i.e., how much resolving the underspecified hypothesis class helps when the model must still find the relevant attributes itself.
\item[\texttt{rule} $\rightarrow$ \texttt{axes\_conj}:] the attribute axes (A) alone, i.e., how much supplying the attribute space helps when the labels of the positive queries remain underdetermined.
\item[\texttt{axes\_conj} $\rightarrow$ \texttt{axes} and \texttt{single} $\rightarrow$ \texttt{axes}:] each piece given the other.
\item[\texttt{axes} $\rightarrow$ \texttt{values}:] the candidate values in addition to the axes (main paper).
\end{description}

Table~\ref{tab:supp-prior-ablation} reports the results for the six models used with HEA, and we relate them to the three observations of the main paper.

\noindent
\textbf{Models struggle to construct the hypothesis space.}
This observation holds beyond the underspecification for the two models with the largest gaps.
Gemma 4 31B gains 0.025 from \texttt{rule} to \texttt{axes\_conj} (0.631 to 0.656) but loses 0.038 to \texttt{single} (0.593), and \texttt{axes} adds only 0.015 on top of \texttt{axes\_conj} (0.671): its gain from \texttt{bare} to \texttt{axes} comes from the attribute axes, not from the single-attribute constraint.
Qwen3-VL 32B gains from either piece alone (0.590 to 0.619 with \texttt{single} and to 0.628 with \texttt{axes\_conj}) and most from both (0.659), so both the hypothesis class and the attribute space are missing when it is not told them.
For both models, naming the attribute space helps even though the hypothesis class still admits conjunctions and the support leaves the positive labels undetermined; the difficulty therefore lies, at least in part, in finding the relevant attributes rather than only in an underspecified task.
For InternVL3.5 38B, most of the gain already comes from \texttt{rule} (0.584 to 0.607), and the four instructions that state a rule differ by at most 0.016; for this model, the task framing matters more than either piece, consistent with its perception rather than reasoning bottleneck (Sec.~\ref{sec:supp-bottleneck}).

\noindent
\textbf{The effect is capability-dependent.}
The smaller models (Qwen3-VL 8B, Gemma 4 E4B, and InternVL3.5 8B) stay between 0.50 and 0.59 under every instruction: neither the attribute space nor the constraint helps them, consistent with the main paper's finding that smaller VLMs benefit little from structural priors.

\noindent
\textbf{The benefit comes from structure rather than more information.}
In the main paper, adding the candidate values to \texttt{axes} does not consistently help.
Here, the axes help Gemma 4 31B and Qwen3-VL 32B even without the single-attribute constraint, while the constraint alone helps only Qwen3-VL 32B.
Both results point to the organization of the hypothesis space along the relevant attributes as the information that matters most.

Finally, the general formulation (Sec.~\ref{sec:supp-hypothesis-space}) allows a support set to determine the labels of the queries without identifying a unique concept.
The synthetic dataset always identifies a single attribute under the single-attribute constraint, so this broader aspect of the formulation is not exercised there; the real dataset, where the concept is a species rather than an attribute, does not impose the constraint.

\begin{table}[t]
\centering
\caption{\textbf{Separating the attribute axes from the single-attribute constraint (Balanced Accuracy).}
\texttt{single}: \texttt{rule} plus the single-attribute constraint, without naming the axes.
\texttt{axes\_conj}: \texttt{rule} plus the attribute axes, with combinations of attributes allowed.
\texttt{axes}: both.
The other columns are taken from Table~\ref{tab:instruction_results}.
All 1,809 synthetic episodes; 95\% bootstrap intervals over target triples are within $\pm$0.011.
Bold denotes the highest score within each model row.}
\label{tab:supp-prior-ablation}
\scriptsize
\setlength{\tabcolsep}{2pt}
\begin{tabular*}{\linewidth}{@{\extracolsep{\fill}}l cccccc@{}}
\toprule
\textbf{Model}
& \texttt{bare}
& \texttt{rule}
& \texttt{single}
& \texttt{axes\_conj}
& \texttt{axes}
& \texttt{values} \\
\midrule
Qwen3-VL-8B~\cite{qwen3-vl}       & 0.570 & 0.569 & \textbf{0.582} & 0.560 & 0.563 & 0.569 \\
Qwen3-VL-32B~\cite{qwen3-vl}      & 0.551 & 0.590 & 0.619 & 0.628 & \textbf{0.659} & 0.652 \\
\hdashline
Gemma 4 E4B~\cite{gemma4}         & 0.520 & 0.518 & \textbf{0.537} & 0.506 & 0.505 & 0.438 \\
Gemma 4 31B~\cite{gemma4}         & 0.605 & 0.631 & 0.593 & 0.656 & 0.671 & \textbf{0.677} \\
\hdashline
InternVL3.5-8B~\cite{internvl35}  & 0.535 & \textbf{0.543} & 0.537 & 0.534 & 0.522 & 0.518 \\
InternVL3.5-38B~\cite{internvl35} & 0.584 & 0.607 & 0.615 & 0.601 & \textbf{0.617} & 0.584 \\
\bottomrule
\end{tabular*}
\end{table}

\subsection{Hypothesize--Eliminate--Apply: Details}
\label{sec:supp-baseline}

HEA (Sec.~\ref{sec:hea}) implements its three stages with five steps (Hypothesize: (1)--(3); Eliminate: (4); Apply: (5)), each a separate call to the same VLM $M$ with greedy decoding.
\textbf{(1) Describe:} $M$ describes each image in $\mathcal{S}$.
\textbf{(2) Discover:} given $\mathcal{S}$ and the descriptions, $M$ lists up to four aspects in which the examples differ.
\textbf{(3) Hypothesize:} $M$ proposes up to five candidate rules, each a statement about the subject of a single image (\eg, ``the animal is covered in sheep wool''); no form of the rule is imposed.
\textbf{(4) Eliminate:} each candidate is checked on each image in $\mathcal{S}$ with a yes/no question; it scores one point if $I^+$ satisfies it and one point for each $I^-_j$ that does not, and the highest-scoring candidate is kept (ties go to the earlier one).
\textbf{(5) Apply:} $I_q$ is classified as positive if it satisfies the kept candidate.

\noindent
\textbf{Natural images.}
Since species are hard to tell apart in words, candidates are given as image crops: step (2) lists parts instead of aspects, step (3) gives bounding boxes of up to eight parts of $I^+$ that may make it positive, and steps (4)--(5) show the crop next to the image and ask whether its subject has the same part.
A whole-subject candidate (``looks like $I^+$ rather than $\{I^-_j\}$'') is also included.
Table~\ref{tab:supp-hea-prompts} lists the main prompts.

\begin{table*}[t]
\centering
\caption{\textbf{Main prompts of HEA.} $\langle\cdot\rangle$ marks inserted content. In steps (2) and (3), the prompt is preceded by the labeled support images and their step-(1) descriptions.}
\label{tab:supp-hea-prompts}
\scriptsize
\begin{tabular}{@{}lp{0.82\textwidth}@{}}
\toprule
\textbf{Step} & \textbf{Prompt} \\
\midrule
(1) Describe & \texttt{Describe the main subject of this image in detail: what it is, and every visible aspect of its appearance. Be specific. Write at most four sentences.} \\
(2) Discover & \texttt{Before working out the rule, list the aspects of the main subject in which these example images differ from one another (up to 4). Give each a short name in capitals and a one-line explanation of what it covers.} (natural: \texttt{parts} instead of \texttt{aspects}) \\
(3) Hypothesize & \texttt{Work out what makes an image POSITIVE. Propose up to 5 candidate rules, most likely first. Phrase each candidate as a statement about the main subject of a single image, so that it can be checked by looking at that image alone.} \\
\quad (natural) & \texttt{Work out what makes an image POSITIVE. Point out up to 8 of these parts in the POSITIVE example (Image 1) that show what makes it POSITIVE, most likely first. For each, name the part and give its bounding box in Image 1.} \\
(4)(5) Check & \texttt{Look at the main subject of this image. Statement: "$\langle$candidate$\rangle$" Is this statement true of this image? Answer with exactly one word: "yes" or "no".} \\
\quad (natural) & \texttt{Image 1 is a close-up of one part ($\langle$part$\rangle$) of an example. Does the main subject of Image 2 have this part, looking the same as in Image 1? Answer with exactly one word: "yes" or "no".} \\
\bottomrule
\end{tabular}
\end{table*}

\section{Bottleneck Analysis}
\label{sec:supp-bottleneck}

To locate where VLMs fail, we decompose concept identification into four stages and evaluate diagnostic conditions in which some stages are performed for the model.
\begin{description}
\item[S1 Perception:] read the shape, texture, and pattern of each image, in $\mathcal{S}$ and of $I_q$.
\item[S2 Hypothesis:] list the candidate attributes in $\mathcal{V}(I^+)$.
\item[S3 Elimination:] reject the candidates shared with any $I^-_j$; for P-negatives, the results of $\text{P}_1$ and $\text{P}_2$ must be combined.
\item[S4 Application:] check whether $I_q$ has the remaining attribute.
\end{description}
All conditions use the same 1,809 episodes and the \texttt{axes} instruction; the \texttt{axes} result on images serves as the reference.
We name the conditions as in the main text: Oracle Attributes (OA) gives the ground-truth attributes of every image, Predicted Attributes (PA) gives the model's own attribute predictions, and Oracle Rule (OR) gives the ground-truth rule together with the ground-truth attributes of the query.
OA-S and OA-Q give the ground-truth attributes only for the support set or only for the query, OR-I gives the rule while the query remains an image, and PA-Solver passes the model's predicted attributes to a deterministic solver.
Table~\ref{tab:supp-conditions} lists which stages each condition requires from the model, and Table~\ref{tab:supp-bottleneck} reports the BA of every condition.
Ground-truth attributes are given as text, e.g., ``shape: horse, texture: sheep wool, pattern: plain''.
In the attribute-recognition probe (AP), the model sees one image and selects its shape, texture, and pattern from the candidate values in Table~\ref{tab:concepts}; PA and PA-Solver use these predictions in place of the ground truth.

\begin{table*}[t]
\centering
\caption{\textbf{Diagnostic conditions.} \checkmark: the model performs the stage itself; --: the stage is not needed or its answer is given; AP: the model's own attribute predictions from the recognition probe AP are used; code: a deterministic solver performs the stage. Perceived \texttt{axes} is the subset of \texttt{axes} queries for which the model reads all three attributes of $I^+$, every $I^-_j$, and $I_q$ correctly in AP.}
\label{tab:supp-conditions}
\small
\begin{tabular}{@{}lll ccccc@{}}
\toprule
& \multicolumn{2}{c}{\textbf{Input}} & \multicolumn{2}{c}{\textbf{S1 Perception}} & \textbf{S2} & \textbf{S3} & \textbf{S4} \\
\cmidrule(lr){2-3} \cmidrule(lr){4-5}
\textbf{Condition} & \textbf{Support $\mathcal{S}$} & \textbf{Query $I_q$} & $\mathcal{S}$ & $I_q$ & \textbf{Hypothesis} & \textbf{Elimination} & \textbf{Application} \\
\midrule
\texttt{axes} & images & image & \checkmark & \checkmark & \checkmark & \checkmark & \checkmark \\
OA-S & ground-truth attributes & image & -- & \checkmark & \checkmark & \checkmark & \checkmark \\
OA-Q & images & ground-truth attributes & \checkmark & -- & \checkmark & \checkmark & \checkmark \\
OA & ground-truth attributes & ground-truth attributes & -- & -- & \checkmark & \checkmark & \checkmark \\
PA & AP attributes & AP attributes & AP & AP & \checkmark & \checkmark & \checkmark \\
OR-I & the rule & image & -- & \checkmark & -- & -- & \checkmark \\
OR & the rule & ground-truth attributes & -- & -- & -- & -- & \checkmark \\
PA-Solver & AP attributes & AP attributes & AP & AP & code & code & code \\
Perceived \texttt{axes} & images (correctly read) & image (correctly read) & \checkmark & \checkmark & \checkmark & \checkmark & \checkmark \\
\bottomrule
\end{tabular}
\end{table*}

\paragraph{Capability gaps.}
Each pair of conditions that differ in exactly one stage isolates one capability.
We define the following gaps in BA:
\begin{align}
\Delta_{\text{reason}} &= \text{OR} - \text{OA}, \label{eq:gap-reason}\\
\Delta_{\text{perc}}   &= \text{OA} - \text{PA}, \label{eq:gap-perc}\\
\Delta_{\text{vis}}    &= \text{PA} - \text{\texttt{axes}}. \label{eq:gap-vis}
\end{align}
$\Delta_{\text{reason}}$ is the cost of inferring the rule (S2--S3) when every attribute is given correctly as text: OR and OA receive the same textual attributes, but only OR is also given the rule.
$\Delta_{\text{perc}}$ is the cost of the model's own misperception (S1): OA and PA differ only in whether the textual attributes are the ground truth or the model's own AP predictions.
$\Delta_{\text{vis}}$ is the cost of reasoning over images rather than over the attributes the model itself reads from them.
The three gaps telescope, so the shortfall of \texttt{axes} from a perfect score splits exactly into
\begin{equation}
1 - \text{\texttt{axes}} = (1 - \text{OR}) + \Delta_{\text{reason}} + \Delta_{\text{perc}} + \Delta_{\text{vis}},
\label{eq:gap-decomp}
\end{equation}
where $1-\text{OR}$ is the failure to apply a given rule to textual attributes.
We further use two gaps outside this chain:
\begin{align}
\Delta_{\text{query}} &= \text{OR} - \text{OR-I}, \label{eq:gap-query}\\
\Delta_{\text{P}}     &= \text{OA}^{\text{F}} - \text{OA}^{\text{P}}, \label{eq:gap-p}
\end{align}
where $\Delta_{\text{query}}$ is the cost of perceiving the query image when the rule is given (S1 on $I_q$ with S4), and $\Delta_{\text{P}}$ compares OA on F-negatives (f1--f5) and P-negatives (p2--p5), i.e., the cost of integrating complementary negatives in S3 when perception is perfect.
Table~\ref{tab:supp-gaps} reports the gaps, computed from Table~\ref{tab:supp-bottleneck}.

\begin{table}[t]
\centering
\caption{\textbf{Capability gaps} (BA, \texttt{axes}; Eqs.~\eqref{eq:gap-reason}--\eqref{eq:gap-p}). The larger of $\Delta_{\text{reason}}$ and $\Delta_{\text{perc}}$ is in bold. Gaps are computed before rounding. $1-\text{OR}$, the remaining term of Eq.~\eqref{eq:gap-decomp}, is 0.000 except for Qwen3-VL 2B (0.024), InternVL3.5 1B (0.325), and InternVL3.5 2B (0.019).}
\label{tab:supp-gaps}
\footnotesize
\setlength{\tabcolsep}{3pt}
\begin{tabular}{@{}lcccccc@{}}
\toprule
\textbf{Model} & \texttt{axes} & $\Delta_{\text{reason}}$ & $\Delta_{\text{perc}}$ & $\Delta_{\text{vis}}$ & $\Delta_{\text{query}}$ & $\Delta_{\text{P}}$ \\
\midrule
\multicolumn{7}{@{}l}{\textit{Qwen3-VL}} \\
\quad 2B & 0.530 & \textbf{0.516} & 0.011 & $-$0.080 & 0.413 & 0.123 \\
\quad 4B & 0.518 & \textbf{0.317} & 0.097 & 0.069 & 0.172 & 0.248 \\
\quad 8B & 0.563 & \textbf{0.390} & 0.038 & 0.009 & 0.179 & 0.128 \\
\quad 30B-A3B & 0.515 & \textbf{0.445} & $-$0.004 & 0.044 & 0.139 & 0.021 \\
\quad 32B & 0.659 & \textbf{0.223} & 0.063 & 0.055 & 0.124 & 0.214 \\
\midrule
\multicolumn{7}{@{}l}{\textit{Gemma 4}} \\
\quad E2B & 0.522 & \textbf{0.467} & 0.014 & $-$0.002 & 0.279 & $-$0.021 \\
\quad E4B & 0.505 & \textbf{0.422} & 0.065 & 0.009 & 0.246 & $-$0.015 \\
\quad 12B & 0.610 & \textbf{0.447} & 0.012 & $-$0.068 & 0.179 & $-$0.031 \\
\quad 31B & 0.671 & \textbf{0.151} & 0.129 & 0.049 & 0.117 & 0.238 \\
\midrule
\multicolumn{7}{@{}l}{\textit{InternVL3.5}} \\
\quad 1B & 0.550 & \textbf{0.099} & 0.048 & $-$0.022 & 0.044 & 0.004 \\
\quad 2B & 0.529 & \textbf{0.306} & 0.158 & $-$0.012 & 0.261 & 0.324 \\
\quad 4B & 0.512 & \textbf{0.309} & 0.154 & 0.025 & 0.258 & 0.308 \\
\quad 8B & 0.522 & 0.176 & \textbf{0.285} & 0.017 & 0.259 & 0.272 \\
\quad 14B & 0.511 & 0.094 & \textbf{0.343} & 0.052 & 0.276 & 0.186 \\
\quad 30B-A3B & 0.527 & 0.218 & \textbf{0.220} & 0.036 & 0.205 & 0.199 \\
\quad 38B & 0.617 & 0.041 & \textbf{0.278} & 0.064 & 0.194 & 0.078 \\
\bottomrule
\end{tabular}
\end{table}

\begin{table*}[t]
\centering
\caption{\textbf{Bottleneck analysis (Balanced Accuracy, \texttt{axes}).} Conditions are defined in Table~\ref{tab:supp-conditions}. For PA-Solver, the percentage is the fraction of episodes in which the solver arrives at a unique rule; the other episodes count as wrong. For Perceived \texttt{axes}, $n$ is the number of queries. AP reports the attribute-recognition accuracy on single images (chance: 0.125 for shape, 0.2 for texture and pattern). Right: \texttt{axes}, OA, and PA on F-negatives (f1--f5) and P-negatives (p2--p5).}
\label{tab:supp-bottleneck}
\setlength{\tabcolsep}{3pt}
\resizebox{\textwidth}{!}{%
\begin{tabular}{@{}l ccccccc cc ccc cccccc@{}}
\toprule
& \multicolumn{7}{c}{\textbf{Diagnostic conditions}} & & & \multicolumn{3}{c}{\textbf{AP accuracy}} & \multicolumn{2}{c}{\textbf{\texttt{axes}}} & \multicolumn{2}{c}{\textbf{OA}} & \multicolumn{2}{c}{\textbf{PA}} \\
\cmidrule(lr){2-8} \cmidrule(lr){11-13} \cmidrule(lr){14-15} \cmidrule(lr){16-17} \cmidrule(lr){18-19}
\textbf{Model} & \texttt{axes} & OA-S & OA-Q & OA & PA & OR-I & OR & \textbf{PA-Solver} & \textbf{Perceived \texttt{axes}} & Sha. & Tex. & Pat. & F & P & F & P & F & P \\
\midrule
\multicolumn{19}{@{}l}{\textit{Qwen3-VL}} \\
2B      & 0.530 & 0.550 & 0.550 & 0.460 & 0.450 & 0.563 & 0.976 & 0.563 {\scriptsize(63\%)} & 0.505 {\scriptsize($n$=630)}  & 0.990 & 0.899 & 0.529 & 0.531 & 0.528 & 0.515 & 0.392 & 0.468 & 0.427 \\
4B      & 0.518 & 0.509 & 0.509 & 0.683 & 0.587 & 0.828 & 1.000 & 0.620 {\scriptsize(70\%)} & 0.518 {\scriptsize($n$=2472)} & 0.995 & 0.871 & 0.796 & 0.521 & 0.513 & 0.794 & 0.545 & 0.629 & 0.533 \\
8B      & 0.563 & 0.552 & 0.503 & 0.610 & 0.572 & 0.821 & 1.000 & 0.573 {\scriptsize(67\%)} & 0.559 {\scriptsize($n$=1633)} & 0.988 & 0.812 & 0.769 & 0.569 & 0.555 & 0.667 & 0.539 & 0.589 & 0.551 \\
30B-A3B & 0.515 & 0.511 & 0.529 & 0.555 & 0.559 & 0.861 & 1.000 & 0.588 {\scriptsize(66\%)} & 0.515 {\scriptsize($n$=2187)} & 0.997 & 0.890 & 0.765 & 0.523 & 0.506 & 0.564 & 0.543 & 0.573 & 0.542 \\
32B     & 0.659 & 0.661 & 0.629 & 0.777 & 0.714 & 0.876 & 1.000 & 0.631 {\scriptsize(70\%)} & 0.676 {\scriptsize($n$=2750)} & 0.990 & 0.908 & 0.792 & 0.683 & 0.628 & 0.872 & 0.658 & 0.769 & 0.645 \\
\midrule
\multicolumn{19}{@{}l}{\textit{Gemma 4}} \\
E2B & 0.522 & 0.558 & 0.515 & 0.533 & 0.520 & 0.721 & 1.000 & 0.352 {\scriptsize(48\%)} & 0.535 {\scriptsize($n$=82)}   & 0.889 & 0.468 & 0.625 & 0.521 & 0.524 & 0.524 & 0.545 & 0.517 & 0.523 \\
E4B & 0.505 & 0.596 & 0.566 & 0.578 & 0.514 & 0.754 & 1.000 & 0.404 {\scriptsize(53\%)} & 0.632 {\scriptsize($n$=299)}  & 0.954 & 0.570 & 0.623 & 0.527 & 0.477 & 0.572 & 0.587 & 0.528 & 0.495 \\
12B & 0.610 & 0.692 & 0.622 & 0.553 & 0.541 & 0.821 & 1.000 & 0.453 {\scriptsize(57\%)} & 0.610 {\scriptsize($n$=587)}  & 0.948 & 0.696 & 0.733 & 0.633 & 0.581 & 0.540 & 0.570 & 0.544 & 0.539 \\
31B & 0.671 & 0.676 & 0.647 & 0.849 & 0.719 & 0.883 & 1.000 & 0.562 {\scriptsize(65\%)} & 0.733 {\scriptsize($n$=1771)} & 0.988 & 0.839 & 0.786 & 0.711 & 0.621 & 0.954 & 0.717 & 0.765 & 0.663 \\
\midrule
\multicolumn{19}{@{}l}{\textit{InternVL3.5}} \\
1B      & 0.550 & 0.523 & 0.506 & 0.576 & 0.528 & 0.631 & 0.675 & 0.334 {\scriptsize(47\%)} & 0.512 {\scriptsize($n$=27)}  & 0.909 & 0.381 & 0.542 & 0.550 & 0.551 & 0.578 & 0.574 & 0.530 & 0.526 \\
2B      & 0.529 & 0.533 & 0.511 & 0.675 & 0.517 & 0.720 & 0.981 & 0.376 {\scriptsize(50\%)} & 0.540 {\scriptsize($n$=151)} & 0.954 & 0.599 & 0.512 & 0.530 & 0.528 & 0.819 & 0.495 & 0.540 & 0.488 \\
4B      & 0.512 & 0.555 & 0.511 & 0.691 & 0.537 & 0.742 & 1.000 & 0.361 {\scriptsize(50\%)} & 0.538 {\scriptsize($n$=188)} & 0.942 & 0.485 & 0.638 & 0.507 & 0.518 & 0.828 & 0.520 & 0.550 & 0.521 \\
8B      & 0.522 & 0.542 & 0.526 & 0.824 & 0.539 & 0.741 & 1.000 & 0.366 {\scriptsize(48\%)} & 0.529 {\scriptsize($n$=154)} & 0.947 & 0.438 & 0.649 & 0.523 & 0.521 & 0.945 & 0.673 & 0.565 & 0.507 \\
14B     & 0.511 & 0.524 & 0.509 & 0.906 & 0.563 & 0.724 & 1.000 & 0.343 {\scriptsize(45\%)} & 0.484 {\scriptsize($n$=65)}  & 0.939 & 0.393 & 0.512 & 0.511 & 0.511 & 0.989 & 0.802 & 0.582 & 0.539 \\
30B-A3B & 0.527 & 0.547 & 0.520 & 0.782 & 0.563 & 0.795 & 1.000 & 0.375 {\scriptsize(51\%)} & 0.559 {\scriptsize($n$=237)} & 0.946 & 0.554 & 0.676 & 0.531 & 0.522 & 0.871 & 0.672 & 0.578 & 0.544 \\
38B     & 0.617 & 0.588 & 0.598 & 0.959 & 0.681 & 0.806 & 1.000 & 0.459 {\scriptsize(57\%)} & 0.617 {\scriptsize($n$=679)} & 0.978 & 0.631 & 0.752 & 0.629 & 0.601 & 0.993 & 0.915 & 0.707 & 0.648 \\
\bottomrule
\end{tabular}}
\end{table*}

\paragraph{Hypothesis formation and elimination ($\Delta_{\text{reason}}$).}
$\Delta_{\text{reason}}$ is the largest term for 11 of the 16 models: all Qwen3-VL models (0.22--0.52), Gemma 4 E2B--12B (0.42--0.47), and InternVL3.5 1B--4B (0.10--0.31; InternVL3.5 1B also fails to apply a given rule, $1-\text{OR}=0.33$).
Even with every attribute given correctly, these models reach only 0.46--0.78 in OA, whereas applying the given rule (OR) is solved almost perfectly.
Restricting \texttt{axes} to the queries whose images they perceive correctly does not help either (e.g., 0.515 $\rightarrow$ 0.515 for Qwen3-VL 30B-A3B and 0.610 $\rightarrow$ 0.610 for Gemma 4 12B; Table~\ref{tab:supp-bottleneck}).
For these models, the failure lies in inferring the rule from the examples, independently of perception.

\begin{figure}
    \centering
    \includegraphics[width=\linewidth]{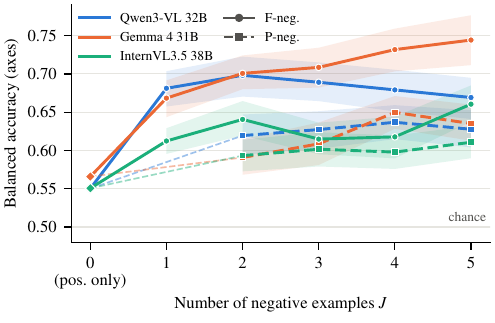}
    \caption{\textbf{Effect of the number of negative examples on the synthetic dataset.} Balanced accuracy under \texttt{axes} for the largest model in each family. Solid and dashed lines denote F- and P-negatives, respectively. $J=0$ is a positive-only reference; shaded regions show 95\% bootstrap intervals.}
    \label{fig:scaling}
\end{figure}

\paragraph{Integrating complementary negatives ($\Delta_{\text{P}}$).}
Within S3, combining P-negatives is the hardest part.
Among the models that infer rules from text (OA$^{\text{F}} \ge 0.79$), $\Delta_{\text{P}}$ is 0.19--0.32: Gemma 4 31B drops from 0.954 to 0.717, Qwen3-VL 32B from 0.872 to 0.658, and InternVL3.5 8B from 0.945 to 0.673; only InternVL3.5 38B keeps a small gap (0.993 to 0.915, $\Delta_{\text{P}}=0.08$).
A solver with ground-truth attributes achieves 1.00 on both categories, so they are equally determined logically.
Models can reject two hypotheses with a single F-negative, but fail to reject different hypotheses with different negatives and combine the results; this explains the F--P gap in the main results (Fig.~\ref{fig:scaling}).
For models with near-chance OA (\eg, Gemma 4 E2B--12B), $\Delta_{\text{P}}$ is close to zero because they fail on both categories.

\paragraph{Perception ($\Delta_{\text{perc}}$ and $\Delta_{\text{query}}$).}
$\Delta_{\text{perc}}$ dominates for the InternVL3.5 models that reason well over text: 0.29, 0.34, and 0.28 for 8B, 14B, and 38B, exceeding their $\Delta_{\text{reason}}$ (0.18, 0.09, and 0.04), and 0.22 for 30B-A3B, on par with $\Delta_{\text{reason}}$.
These models solve the episodes from correct text (OA of 0.78--0.96) but fall to 0.54--0.68 with their own readings (PA), consistent with their low texture accuracy in AP (0.39--0.63).
Because one decision involves three to seven images ($I^+$, $J$ negatives, and $I_q$), misreading any one of them can break the elimination.
For Gemma 4 31B, $\Delta_{\text{perc}}$ (0.13) is comparable to $\Delta_{\text{reason}}$ (0.15); for Qwen3-VL 32B, it is smaller (0.06 vs.\ 0.22).
$\Delta_{\text{query}}$ shows that perceiving the query alone already costs 0.12--0.28 for 14 of the 16 models, even though the rule is given (OR-I of 0.72--0.88).
Perception is thus a secondary bottleneck for most models and the primary one for the larger InternVL3.5 models.

\paragraph{Reasoning over images ($\Delta_{\text{vis}}$).}
$\Delta_{\text{vis}}$ is small for every model ($-$0.08 to +0.07).
Once a model's own readings are fixed, presenting the episode as images rather than as those readings changes little; the loss from images comes from misreading them ($\Delta_{\text{perc}}$), not from reasoning over the visual input itself.

\paragraph{A modular pipeline.}
If the model only reads attributes and a solver performs S2--S4 (PA-Solver), $\Delta_{\text{reason}}$ is removed, but perception errors remain: the solver arrives at a unique rule in only 45--70\% of episodes, and PA-Solver reaches 0.33--0.63 BA when the remaining episodes count as wrong, below \texttt{axes} for 12 of the 16 models (Table~\ref{tab:supp-bottleneck}).

\paragraph{What the two gaps measure.}
The two gaps are measured asymmetrically: $\Delta_{\text{reason}}$ is the cost of reasoning with perfect perception (OA vs.\ OR), whereas $\Delta_{\text{perc}}$ is the cost of misperception through the model's own reasoning (PA vs.\ OA), so a model that cannot reason shows a small $\Delta_{\text{perc}}$ even if it misreads images.
The main paper's statement that Qwen3-VL and Gemma 4 have larger reasoning gaps refers to these quantities.
To examine this conclusion in more detail, we start from PA and fix one capability at a time (Table~\ref{tab:supp-gap-orders}): perception, by giving ground-truth attributes (OA), or reasoning, by passing the model's readings to a solver (PA-Solver$^{+}$, a variant of PA-Solver that never abstains and guesses randomly when the kept axes disagree).
The capability whose fix helps more is the larger bottleneck.
This symmetric comparison confirms the main paper's conclusion for 13 of the 16 models and refines it at two points: Gemma 4 31B, the strongest Gemma 4 model, gains more from correct attributes (0.849) than from perfect reasoning (0.793), so it is limited by perception at least as much as by reasoning; and InternVL3.5 shifts to a perception bottleneck already at 2B rather than at 8B.

Two caveats apply to all of these comparisons.
First, changing the input format is itself an intervention: Qwen3-VL 2B and Gemma 4 12B score lower with ground-truth attributes given as text (OA, 0.460 and 0.553) than with the images (\texttt{axes}, 0.530 and 0.610), so OA does not upper-bound what a model can do from images.
Second, AP reads each image with a separate recognition prompt, and these readings need not match the attributes the model relies on when it classifies directly.
The conclusion that holds regardless of these choices is that many models fail even on the textual version of the task (OA of 0.46--0.69 for ten models), although all but InternVL3.5 1B apply a given rule almost perfectly (OR $\geq$ 0.98).

\begin{table}[t]
\centering
\caption{\textbf{Which fix helps more?} (BA, \texttt{axes}). PA: the model reads the images and reasons. Fix perception (OA): the model reasons over ground-truth attributes. Fix reasoning (PA-Solver$^{+}$): the model's own readings are passed to a solver with perfect reasoning. The higher of the two fixes is in bold, and the capability it fixes is the larger bottleneck.}
\label{tab:supp-gap-orders}
\footnotesize
\setlength{\tabcolsep}{4pt}
\begin{tabular}{@{}lcccc@{}}
\toprule
\textbf{Model} & PA & Fix perception & Fix reasoning & Bottleneck \\
\midrule
\multicolumn{5}{@{}l}{\textit{Qwen3-VL}} \\
\quad 2B & 0.450 & 0.460 & \textbf{0.780} & Reasoning \\
\quad 4B & 0.587 & 0.683 & \textbf{0.820} & Reasoning \\
\quad 8B & 0.572 & 0.610 & \textbf{0.785} & Reasoning \\
\quad 30B-A3B & 0.559 & 0.555 & \textbf{0.811} & Reasoning \\
\quad 32B & 0.714 & 0.777 & \textbf{0.828} & Reasoning \\
\midrule
\multicolumn{5}{@{}l}{\textit{Gemma 4}} \\
\quad E2B & 0.520 & 0.533 & \textbf{0.650} & Reasoning \\
\quad E4B & 0.514 & 0.578 & \textbf{0.672} & Reasoning \\
\quad 12B & 0.541 & 0.553 & \textbf{0.721} & Reasoning \\
\quad 31B & 0.719 & \textbf{0.849} & 0.793 & Perception \\
\midrule
\multicolumn{5}{@{}l}{\textit{InternVL3.5}} \\
\quad 1B & 0.528 & 0.576 & \textbf{0.632} & Reasoning \\
\quad 2B & 0.517 & \textbf{0.675} & 0.667 & Perception \\
\quad 4B & 0.537 & \textbf{0.691} & 0.647 & Perception \\
\quad 8B & 0.539 & \textbf{0.824} & 0.650 & Perception \\
\quad 14B & 0.563 & \textbf{0.906} & 0.638 & Perception \\
\quad 30B-A3B & 0.563 & \textbf{0.782} & 0.663 & Perception \\
\quad 38B & 0.681 & \textbf{0.959} & 0.718 & Perception \\
\bottomrule
\end{tabular}
\end{table}

\paragraph{Summary.}
With perception made perfect, reasoning remains the larger cost for Qwen3-VL and for Gemma 4 up to 12B: these models fail on the textual version of the task although they apply a given rule almost perfectly, so forming and eliminating hypotheses from the support set is the main difficulty, and within it the integration of complementary P-negatives ($\Delta_{\text{P}}$).
For InternVL3.5 at 8B and larger, perception is the larger cost under every measure.
Gemma 4 31B lies between the two: its reasoning gap in the main paper's decomposition is larger, but it gains more from correct attributes than from perfect reasoning (Table~\ref{tab:supp-gap-orders}).
Reasoning over images rather than over the attributes the model reads from them ($\Delta_{\text{vis}}$) costs little throughout.

\section{Positive-only and Negative-only References}
\label{sec:supp-posneg}

To check that both kinds of support examples are needed, we evaluate each VLM on the Synthetic Gavagai Dataset with only the positive example ($\mathcal{S}=\{I^+\}$) or only the negative examples ($\mathcal{S}=\{I^-_j\}$), on the same queries.
Table~\ref{tab:supp-posneg} reports the results.
VLMs score at most 0.58 with either kind alone, and improve only when both are given, as the concept is determined only by eliminating hypotheses with the negatives.

\begin{table}[t]
\centering
\caption{\textbf{Positive-only and negative-only references on the synthetic dataset (BA).} P+N: both kinds of support examples; P: the positive only; N: the negatives only. All models use the \texttt{axes} instruction.}
\label{tab:supp-posneg}
\footnotesize
\begin{tabular}{@{}l ccc@{}}
\toprule
\textbf{Model} & P+N & P & N \\
\midrule
\multicolumn{4}{@{}l}{\textit{Qwen3-VL}} \\
\quad 2B & 0.530 & 0.502 & 0.477 \\
\quad 4B & 0.518 & 0.502 & 0.462 \\
\quad 8B & 0.563 & 0.527 & 0.492 \\
\quad 30B-A3B & 0.515 & 0.499 & 0.507 \\
\quad 32B & 0.659 & 0.550 & 0.523 \\
\midrule
\multicolumn{4}{@{}l}{\textit{Gemma 4}} \\
\quad E2B & 0.522 & 0.493 & 0.491 \\
\quad E4B & 0.505 & 0.503 & 0.375 \\
\quad 12B & 0.610 & 0.548 & 0.528 \\
\quad 31B & 0.671 & 0.566 & 0.576 \\
\midrule
\multicolumn{4}{@{}l}{\textit{InternVL3.5}} \\
\quad 1B & 0.550 & 0.517 & 0.497 \\
\quad 2B & 0.529 & 0.498 & 0.508 \\
\quad 4B & 0.512 & 0.511 & 0.493 \\
\quad 8B & 0.522 & 0.509 & 0.499 \\
\quad 14B & 0.511 & 0.520 & 0.471 \\
\quad 30B-A3B & 0.527 & 0.514 & 0.527 \\
\quad 38B & 0.617 & 0.551 & 0.536 \\
\bottomrule
\end{tabular}
\end{table}

\section{Dataset Examples}
\label{sec:supp-examples}

\subsection{Synthetic Gavagai Dataset}
\begin{figure*}[p]
    \centering
    \includegraphics[width=\linewidth]{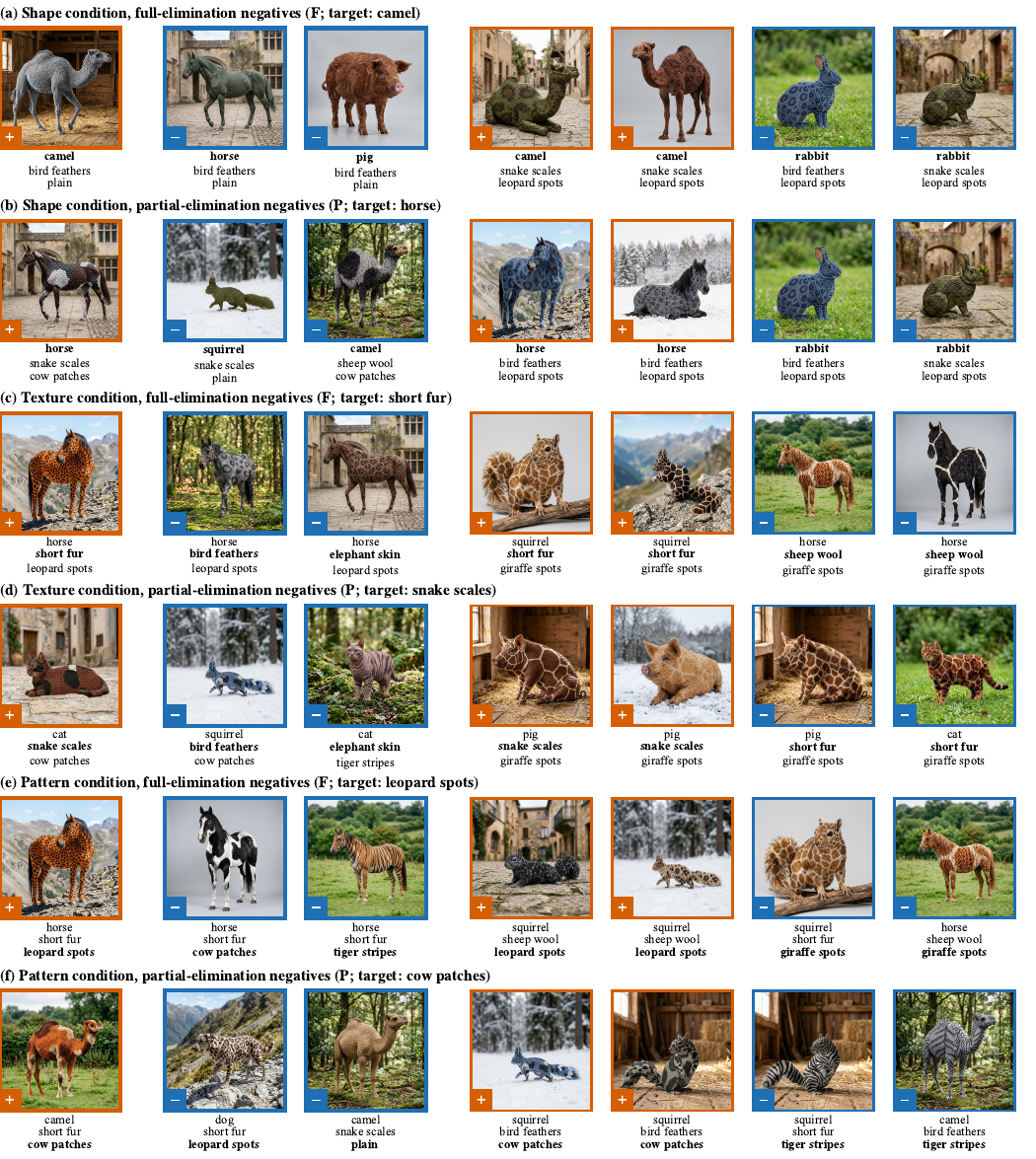}
    \caption{\textbf{Examples from the Synthetic Gavagai Dataset.} One f2 (F) and one p2 (P) episode per target condition. In each row, the left three images are the support set (the positive example $I^+$ and the two negatives $I^-_j$) and the right four are queries (two positive and two negative of the eight); orange/$+$ marks positive images and blue/$-$ negative ones. Below each image are its shape, texture, and pattern; the target attribute is in bold.}
    \label{fig:supp-examples-synthetic}
\end{figure*}

Figure~\ref{fig:supp-examples-synthetic} shows two episodes for each target condition, one with full-elimination (F) and one with partial-elimination (P) negatives.
An F-negative differs from the positive example only in the target attribute, so a single negative rules out both non-target attributes.
In (a), for example, the horse and the pig share the bird-feather texture and the plain pattern with the positive camel, leaving the camel shape as the concept.
A P-negative instead shares one non-target attribute with the positive example and rules out only that attribute, so the negatives must be combined.
In (b), the squirrel shares the snake-scale texture with the positive horse and the camel shares its cow-patch pattern; only together do they leave the horse shape.
In both rows, queries with the target shape are positive whatever their texture and pattern, while the rabbits are negative even when they share the texture or pattern of the positive queries.

\subsection{Real Gavagai Dataset}
\begin{figure*}[p]
    \centering
    \includegraphics[width=\linewidth]{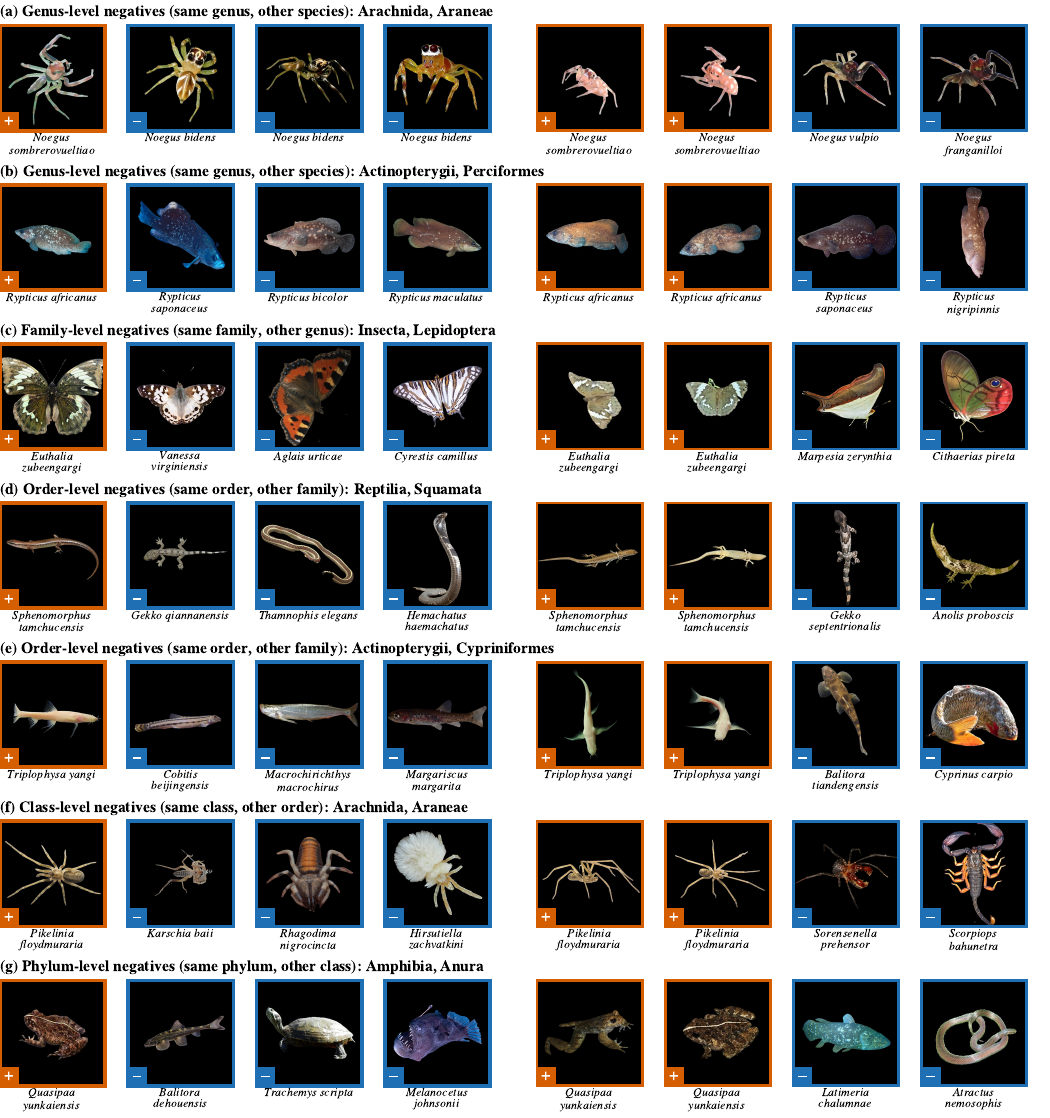}
    \caption{\textbf{Examples from the Real Gavagai Dataset.} Seven episodes from different classes, with negatives at the rank given in each row title; the negatives at rank $r$ share rank $r$ with $I^+$ but differ at the next lower rank. In each row, the left four images are the support set (the positive example $I^+$ and the $J=3$ negatives $I^-_j$) and the right four are queries (two positive and two negative of the six); orange/$+$ marks positive images and blue/$-$ negative ones. Below each image is its species.}
    \label{fig:supp-examples-real}
\end{figure*}

Figure~\ref{fig:supp-examples-real} shows seven episodes from different animal classes, with negatives at every taxonomic rank, as they are given to the models.
The positive queries are other individuals or other views of the novel species, not copies of $I^+$.
At the class and phylum ranks (f, g), the negatives differ from the positive in body plan, e.g., two solifuges and a mite for the positive spider and a loach, a turtle, and an anglerfish for the positive frog, so a coarse visual category suffices to reject the negative queries.
At the order rank (d), the negatives are a gecko and snakes, which share the order Squamata with the positive skink, while the negative queries are other lizards, so the model must go beyond the lizard body form.
At the genus rank (a, b), the negatives and negative queries are congeneric jumping spiders or fishes that share the body form and much of the coloration of the positive, and recognizing the novel species requires attending to subtle differences in shape and markings.

\end{document}